\documentclass[11pt,a4paper]{article}

\usepackage[T1]{fontenc}
\usepackage{lmodern}
\usepackage[a4paper,textwidth=165mm,top=24mm,bottom=25mm]{geometry}
\usepackage{microtype}
\usepackage{amsmath,amssymb}
\usepackage{graphicx}
\usepackage{booktabs,tabularx}
\usepackage{float}
\usepackage[numbers,sort&compress]{natbib}
\usepackage{xcolor}
\usepackage[colorlinks=true,linkcolor=blue!45!black,citecolor=blue!45!black,urlcolor=blue!45!black]{hyperref}

\title{\bfseries Inflated, Deflated, Model-Specific: What a Worst-Group Gain from Prompt Optimization Measures}
\author{%
  Andrey Marchenko\textsuperscript{1} \quad Oleg Somov\textsuperscript{1,2} \\
  \textsuperscript{1}Moscow Institute of Physics and Technology \quad \textsuperscript{2}AIRI \\
  \small\href{mailto:marchenko.ai@phystech.edu}{marchenko.ai@phystech.edu}}
\date{}
\hypersetup{
  pdftitle={Inflated, Deflated, Model-Specific: What a Worst-Group Gain from Prompt Optimization Measures},
  pdfauthor={Andrey Marchenko, Oleg Somov},
  pdfkeywords={prompt optimization, group robustness, worst-group accuracy, selection bias, LLM evaluation, cross-model transfer}
}

\begin{document}
\maketitle

\begin{abstract}
Prompt optimization for robustness across subgroups is often evaluated by the worst-group gain of the best prompt. This statistic selects both the best prompt and the lowest-scoring group from noisy estimates. We analyze what it measures on CivilComments, MultiNLI, and multilingual toxicity, comparing a seed prompt with optimizer-selected prompts and one-line strictness controls. We use a paired bootstrap to estimate the best gain expected by chance and track performance on the initially worst group separately. On CivilComments, the best of 36 prompts gains 1.9 percentage points, below the 2.7 points expected under the selection null. Prompts shift predictions, but no prompt significantly improves whole-test balanced accuracy. On multilingual toxicity, the initially worst language gains 6.1 points, while the worst-group gain is not significant because the minimum moves to another language. On MultiNLI, repartitioning the same examples into ten random groups reduces the number of significant gains from 17 to zero at a fixed test budget. None of four significant group-specific gains on Gemma~3~12B replicates on either of two alternative models, and all eight cross-model effect differences are significant. We provide seven checks for interpreting reported gains, six of which use only existing predictions.

\end{abstract}

\section{Introduction}
\label{sec:intro}

Prompt optimization has become a common step in building LLM applications. Practitioners write an
initial instruction, the \emph{seed prompt}, and let an optimizer propose candidates, score them on
labelled data and keep the best one, which we call the \emph{final prompt}, or \emph{final} for short. APE~\citep{zhou2023large}, ProTeGi~\citep{pryzant2023automatic},
EvoPrompt~\citep{guo2024evoprompt}, GPO~\citep{li2023robust} and GEPA~\citep{agrawal2026gepa}
search in very different ways, and frameworks such as DSPy ship several of
them~\citep{khattab2024dspy}, but all are judged by the gain their best prompt shows over the seed.
When the goal is robustness across subgroups, claims of improvement rest on a single number: the
worst-group accuracy of the selected prompt minus that of the seed. Robustness across groups is a
multi-objective problem with one accuracy per group, and this number reduces it to a single scalar,
the minimum. This paper asks what that number measures.

This number rests on two selections: the best of many candidates is picked, and the metric takes
the worst of several groups (Section~\ref{sec:setup} writes both out). Both selections act on noisy
estimates: the accuracy of a prompt measured on a finite test set would come out differently on
another sample of the same size, and we call this variation \emph{noise}. A difference is
\emph{significant} when it survives a correction for multiple comparisons
(Section~\ref{sec:experiments}). The worst group under the seed, the \emph{initially worst group},
does not depend on the search, so the gain on it is measured on the same group for every prompt.

We reproduce such a report on three benchmarks with groups of three kinds: identities mentioned in
comments for toxicity detection on CivilComments~\citep{borkan2019nuanced,koh2021wilds}, genres for
contradiction detection on MultiNLI~\citep{williams2018broad}, and languages for multilingual
toxicity detection~\citep{dementieva2024overview}. The seed is compared with the final prompts of
several optimizers and with edits of the seed that only change how strict it is. Every prompt is run
by one LLM, the \emph{task model}, and the optimizers select prompts by its answers.

A reader takes the number as the improvement that a method brings to the group that fares worst, and
read this way it can err in both directions. Picking the best prompt can yield an apparent gain
within the range of chance, and the minimum can make this inflation larger. When a prompt lifts the
worst group past another, the minimum moves to that group and can hide part of the gain. A gain that remains once
both effects are accounted for can depend on the task model.

\textbf{The winner can be indistinguishable from noise.} The largest of $N$ noisy gains is positive
on average even when no prompt beats the seed, so a reported gain has to be compared with the \emph{chance baseline}: the average gain of the best of $N$ prompts when none of them is truly better than the seed. On CivilComments the best of 36
prompts gains $+0.019$, less than the $+0.027$ that chance alone gives the best of 36 on average
(Figure~\ref{fig:winners-curse}). The prompts do change the model, as worst-class accuracy on the
same predictions shows, but no prompt significantly improves balanced accuracy on the whole test.
The reported worst-group gain is consistent with chance after selection. On MultiNLI the gain passes the same test
(Section~\ref{sec:winner}).

\textbf{The minimum can hide a significant gain.} When a prompt lifts the worst group past another,
the minimum moves and records only part of the improvement. On multilingual toxicity one prompt
raises balanced accuracy on Amharic, the initially worst group, by a significant $+0.061$,
but German slips within noise and becomes the minimum. The worst-group gain for this prompt is
only $+0.024$ and is not significant. The gain on the initially worst group exceeds what a
change of strictness alone achieves (Section~\ref{sec:moves}).

\textbf{The minimum adds noise of its own.} Taking the worst group is a second selection, made on
groups that each hold only part of the test. On all three benchmarks chance alone gives the best of
$N$ prompts about three times the gain by the minimum that it gives by balanced accuracy on the whole
test, and findings of the minimum replicate less often than those of averaged statistics. Where the
worst group changes from one sample to the next, as on CivilComments, the chance baseline rises
further. The number of groups comes from how the benchmark was built, and at a fixed test budget it
sets the resolution of the test. Splitting the same MultiNLI examples into ten random groups widens
the intervals and turns 17 significant gains into none while the effect stays the same
(Section~\ref{sec:stable}).

\textbf{Surviving gains did not transfer.} On our main task model, four finals have a significant
gain on the initially worst group. We ran them unchanged on two other task models from other
developers, which under the seed are about as accurate as the main one and largely agree with it on
which groups are hard. None of the four gains holds on either model. Two distinct prompts turn into
significant losses: APE on both alternative models and GEPA on GPT-4o-mini. For every final the effect
on each other model differs significantly from the gain on the main one. Selection on the main model
would pull a gain back toward zero but would not turn it
into a significant loss, so the reversals point to prompts fitted to the model on which they were
found (Section~\ref{sec:model}).

Our contribution is to measure what worst-group metrics add to the selection of the best prompt: a
second selection inside the metric with noise of its own, an error of the opposite sign, and
dependence on the task model. To assess the selected gain we use the chance baseline, the
Reality Check~\citep{white2000reality} applied to worst-group gains.
Section~\ref{sec:checks} turns these findings into seven checks for a reported worst-group gain, six
of which need only the predictions at hand, and reports five preregistered follow-up studies, failed
predictions included.

\subsection{Related Work}
\label{sec:related}

\paragraph{Prompt optimization.}
The optimizers named in Section~\ref{sec:intro} differ mainly in how they propose candidates; GPO
among them targets distribution shift, and group robustness has also been pursued by tuning
continuous prompts~\citep{phan2024controllable}. These works compare methods by the gain of the
selected prompt on benchmark tasks. Our pools contain final prompts of APE, ProTeGi, GPO, both
EvoPrompt variants, GEPA and PRIME (Appendix~\ref{app:prime}), and we ask what such a gain measures
when the target is the worst group.

\paragraph{Worst-group evaluation.}
Worst-group accuracy became the standard target for robustness to spurious correlations with
group DRO~\citep{sagawa2020distributionally} and the WILDS benchmark, and CivilComments was
released together with subgroup metrics for unintended bias. \citet{yang2023change} find that
worst-class accuracy selects models well even without group labels and that worst-group accuracy
trades off against other metrics; we use worst-class accuracy as a control on the same predictions.
\citet{ethayarajh2020classifier} showed that common datasets are too small to confirm subgroup
bias except in the coarsest cases, and \citet{card2020little} that many NLP experiments lack the
power to detect the differences they test. We start from the same concern. Even with intervals of
the correct width, the number of significant findings depends on the number of groups and on which
test sample is drawn, and the minimum can deflate a real gain of the group it is meant to protect.

\paragraph{Scalarizing per-group objectives.}
Taking the worst of several per-group objectives is the max--min scalarization of a multi-objective
problem~\citep{miettinen1999nonlinear}, and minimax group fairness has been cast as the search for a
Pareto-efficient model with the smallest worst-group risk~\citep{martinez2020minimax}. There the
minimum is the objective being optimized; in prompt optimization it is also the number by which
gains are reported. We study this second use: the minimum records the change of one group and hides
how the others moved, and with noisy per-group estimates both the identity of the minimum and the
number of objectives become sources of error.

\paragraph{Selection and the winner's curse.}
Reporting the best of many compared candidates is a multiple-comparison
problem~\citep{jensen2000multiple}, and the variance of the selection criterion alone can cause
over-fitting in model selection~\citep{cawley2010overfitting}. White's Reality
Check tests whether the best of many models beats a benchmark given that it
was selected, by bootstrapping the maximum while preserving the dependence between candidates;
studentized and stepwise refinements followed~\citep{hansen2005test,romano2005stepwise}, building on
resampling-based max-$T$ adjustments~\citep{westfall1993resampling}. Our chance baseline is this
test applied to worst-group gains. For adaptive prompt and program search,
SIREN~\citep{xu2026towards} proposes a general selection-aware reporting protocol. Worst-group
metrics add two selections that such protocols do not model, namely which group is the minimum and
how many groups are defined, and we measure the effect of each. Choices of model selection are also
known to change conclusions in domain generalization~\citep{gulrajani2021search};
Appendix~\ref{app:protocol} quantifies the analogous protocol choices for prompt optimization.

\paragraph{Prompt sensitivity and transfer.}
LLM performance varies widely with prompt formatting~\citep{sclar2024quantifying} and instruction
templates, which can change even the ranking of models~\citep{mizrahi2024state}, and variance from
data sampling and random seeds affects benchmark conclusions more
generally~\citep{bouthillier2021accounting,agarwal2021deep}. \citet{cao2024worst} study the worst
case over paraphrases of a request; our worst case is over groups of data under a fixed prompt,
and it is this choice of groups that creates the selections above. Discrete prompts induced on one
language model can lose accuracy on another~\citep{rakotonirina2023can}; we find the same for every
worst-group gain that survives our checks.

\section{Materials and Methods}
\label{sec:setup}

We first state the problem in general terms and then describe the experiments.

\subsection{Problem formulation}
\label{sec:formulation}

\paragraph{Entities.}
A benchmark is a binary classification task with examples $(x, y, g)$, a label $y \in \{0,1\}$ and a
group $g \in \mathcal{G}$, $|\mathcal{G}| = K$. A \emph{task model} $M$ executes a prompt $S$ on an
input and returns a label $M(S, x)$. An \emph{optimizer} $\mathcal{O}$ starts from a seed prompt
$S_0$ and returns a final prompt
\[
  S^{*} = \mathcal{O}\bigl(S_0;\, M, M_{\mathrm{opt}}, D_{\mathrm{tr}}, D_{\mathrm{val}}, r\bigr),
\]
where $M_{\mathrm{opt}}$ is the LLM that writes candidate prompts, $D_{\mathrm{tr}}$ and
$D_{\mathrm{val}}$ are the labelled data the search uses, and $r$ is a random seed. We treat
$\mathcal{O}$ as a black box and use only its input and output.

\paragraph{The optimization problem.}
For a prompt $S$, let $\mathrm{BA}_g(S) = \tfrac12[\mathrm{TPR}_g(S) + \mathrm{TNR}_g(S)]$ be the
balanced accuracy of $M$ on group $g$, the mean of its accuracies on the two classes, and let
\[
  W(S) = \min_{g \in \mathcal{G}} \mathrm{BA}_g(S)
\]
be its \emph{worst-group balanced accuracy}. Averaging the two classes within a group before taking
the minimum counts a gain on one class only net of the loss on the other. The minimum over
group--label cells used by WILDS sees only the worse cell, so a prompt can raise it at a large cost
to the other class, and on CivilComments it mostly tracks how often the model says ``toxic''
(Appendix~\ref{app:ceiling}). An optimizer aimed at robustness seeks
$S^{*}$ with $W(S^{*})$ as large as possible and is judged by the gain
$\Delta(S^{*}) = W(S^{*}) - W(S_0)$.

\paragraph{What is reported.}
$W$ is estimated on a finite test set $D_{\mathrm{te}}$, which gives $\widehat{\mathrm{BA}}_g$,
$\widehat W$ and $\widehat\Delta$. A comparison of methods runs several optimizers and seeds,
collects the prompts in a pool $\mathcal{P} = \{S_1, \dots, S_N\}$ and reports the best of them:
\[
  \widehat\Delta_{\max} = \max_{i \le N}
  \Bigl[\min_{g} \widehat{\mathrm{BA}}_g(S_i) - \min_{g} \widehat{\mathrm{BA}}_g(S_0)\Bigr].
\]
Two selections are built into this number, the maximum over prompts and the minimum over groups,
and both act on noisy estimates. The optimizer's own choice of $S^{*}$ on $D_{\mathrm{val}}$ is a
third selection, but it does not use $D_{\mathrm{te}}$, and we test it separately
(Section~\ref{sec:winner}). We ask what $\widehat\Delta_{\max}$ measures as a result.

\paragraph{What we compare it with.}
Each finding sets $\widehat\Delta$ against a quantity that removes one selection or one source of
noise.
\begin{itemize}
  \item \emph{Chance.} Suppose that no prompt is better than the seed, $\Delta(S_i) = 0$ for all
    $i$. Noise alone still gives the best of $N$ a positive gain, $B(N) = \mathbb{E}_0[\widehat\Delta_{\max}]$,
    the \emph{chance baseline}. Its 95th percentile $C_{95}(N)$ is the \emph{noise ceiling}, and the
    Reality Check $p$-value is the probability, under this null, of a best gain at least as large as
    the observed one (Section~\ref{sec:winner}).
  \item \emph{No groups.} With $K = 1$, $W$ becomes the balanced accuracy of the whole test set,
    and comparing the two isolates the effect of grouping. Worst-class accuracy
    $\mathrm{WC}(S) = \min(\mathrm{TPR}(S), \mathrm{TNR}(S))$ over the whole test set shows whether
    $S$ shifts the balance between the two classes (Section~\ref{sec:winner}).
  \item \emph{A fixed group.} The initially worst group $g_0 = \arg\min_g \widehat{\mathrm{BA}}_g(S_0)$,
    the worst group under the seed, does not depend on the search, and the gain on it is
    $\widehat\Delta_{g_0}(S) = \widehat{\mathrm{BA}}_{g_0}(S) - \widehat{\mathrm{BA}}_{g_0}(S_0)$.
    Since $\widehat W(S_0) = \widehat{\mathrm{BA}}_{g_0}(S_0)$ and
    $\widehat W(S) \le \widehat{\mathrm{BA}}_{g_0}(S)$,
    \[
      \widehat\Delta(S) \le \widehat\Delta_{g_0}(S),
    \]
    with equality exactly when $g_0$ remains the worst group under $S$ (Section~\ref{sec:moves}).
  \item \emph{Smoothed minima and the number of groups.} CVaR@25\%, the mean over the worst quarter
    of groups, and a softmin, a smooth version of the minimum taken after shrinking each group
    toward the mean (Appendix~\ref{app:ceiling}), keep the focus on weak groups with less noise. We
    also change $K$ by regrouping the same examples (Section~\ref{sec:stable}).
  \item \emph{Another task model.} $\widehat\Delta_{g_0}(S; M')$ for a model $M' \neq M$
    (Section~\ref{sec:model}).
\end{itemize}

\subsection{Experimental setup}
\label{sec:experiments}

\paragraph{Benchmarks.}
We use three benchmarks of this kind (Table~\ref{tab:benchmarks}). CivilComments, in its WILDS
release, asks whether an online comment is toxic; its groups are eight identities that a comment may
mention, from gender and religion to race, and each comment is assigned to the first identity it
mentions (Appendix~\ref{app:benchmarks}). MultiNLI asks whether a hypothesis contradicts its
premise; its groups are ten genres, from fiction to transcribed speech. Multilingual toxicity asks
the CivilComments question in six languages, English, German, Russian, Arabic, Hindi and Amharic,
and uses the same seed prompt. In every test set each group holds the same number of examples of
each label, so no group is measured less precisely than another. The instability of the worst
group in Section~\ref{sec:stable} therefore does not come from small groups, and the test can be
divided into other groups without changing anything else. The optimizers work on separate data:
$D_{\mathrm{tr}}$ is a small labelled training set, $D_{\mathrm{val}}$ holds validation data, and
neither overlaps $D_{\mathrm{te}}$. On CivilComments and MultiNLI, $D_{\mathrm{tr}}$ comes from only
some of the groups and $g_0$ is not among them, while $D_{\mathrm{val}}$ covers all groups.
Appendix~\ref{app:benchmarks} describes how the sets were built.

\begin{table}[H]
  \caption{\textbf{Benchmarks.} A cell is a group--label pair. Test examples are counted after
    dropping those on which some prompt gave an unparsable answer, so that all prompts are compared
    on the same examples (Appendix~\ref{app:benchmarks}); on CivilComments they include 400 comments
    that mention no identity and count only toward the metrics without groups. The pool holds every prompt
    compared with the seed. The last two rows refer to the seed prompt on the main task model; BA is
    balanced accuracy.}
  \label{tab:benchmarks}
  \centering
  \small
  \setlength{\tabcolsep}{3.5pt}
  \begin{tabular}{@{}lccc@{}}
    \toprule
     & \begin{tabular}[b]{@{}c@{}}Civil-\\Comments\end{tabular}
     & MultiNLI
     & \begin{tabular}[b]{@{}c@{}}Multilingual\\toxicity\end{tabular} \\
    \midrule
    Groups                    & 8 identities & 10 genres & 6 languages \\
    Examples per cell         & 200 & 250 & 250 \\
    Test examples used        & 3,584 & 4,991 & 2,907 \\
    Optimizer finals          & 25 & 22 & 9 \\
    One-line edits            & 11 & 11 & 11 \\
    Pool size $N$             & 36 & 33 & 20 \\
    \midrule
    Seed worst-group BA       & 0.665 & 0.838 & 0.721 \\
    Seed worst group          & white & verbatim & Amharic \\
    \bottomrule
  \end{tabular}
\end{table}

\paragraph{Prompts.}
Our pool $\mathcal{P}$ holds prompts of two kinds. Most are finals of optimizer runs. Inside
$\mathcal{O}$, $M_{\mathrm{opt}}$ writes candidate prompts from labelled examples, from the errors
of the current prompts or by recombining them; $M$ runs each candidate on labelled examples; and
the search continues from the best candidates until its budget is spent. The final is chosen on
$D_{\mathrm{val}}$ by a statistic fixed in the run's configuration. In all our runs
$M_{\mathrm{opt}}$ is DeepSeek-V4-Pro. We ran APE, ProTeGi, GPO, EvoPrompt, GEPA and PRIME, an
evolutionary search that also chooses which training examples it learns from
(Appendix~\ref{app:prime}), on CivilComments, and APE, EvoPrompt and GEPA, plus ProTeGi on toxicity,
on the other two benchmarks, most of them with three random seeds $r$. Finals that duplicate $S_0$
or another prompt are removed; implementations and budgets are given in Appendix~\ref{app:pool}.
The other kind are eleven one-line edits of $S_0$, each replacing its rule for unsure cases with
one of a different strictness. Each changes only that rule, and so mainly how readily the model
says ``yes''; together they show what a shift of the decision threshold alone achieves
(Section~\ref{sec:moves}). They are part of the pool as well, and leaving them out changes no
verdict (Section~\ref{sec:winner}). Appendix~\ref{app:pool} lists them.

\paragraph{Task models.}
The main task model $M$ is Gemma~3 12B~\citep{gemma2025gemma3}. It runs every candidate inside the
searches and every prompt of every pool, so all selections in this paper are made on its answers.
We query the task models through OpenRouter at temperature 0 with at most 16 output tokens.
The parser looks for an explicit label first, then a label word, then a bare digit. We exclude
unparsable examples from the corresponding paired comparisons as detailed in
Appendix~\ref{app:benchmarks}. Evaluation uses human-annotated gold labels. The two models $M'$ come from
other developers: the closed GPT-4o-mini and the open-weight Qwen3-235B. With the seed prompt both are
about as accurate as $M$ (Appendix~\ref{app:second-model}), and Qwen3-235B was
admitted by an accuracy rule fixed before its runs (Appendix~\ref{app:second-model}). Besides the
transfer test of Section~\ref{sec:model}, they replicate results of Sections~\ref{sec:winner}
and~\ref{sec:stable}. Providers, decoding settings and query dates are listed in
Appendix~\ref{app:benchmarks}.

\paragraph{Estimation.}
We call the difference of any of these quantities between a prompt and $S_0$ on the same test
examples a \emph{contrast}. Because both prompts are scored on the same examples, the variation from
easy and hard examples partly cancels, and we estimate the remaining uncertainty with a paired
bootstrap~\citep{dror2018hitchhiker,miller2024adding}: 4,000 resamples drawn within group--label
cells, the same for all prompts. We report 95\% percentile intervals and two-sided $p$-values.
Contrasts that share $S_0$ are corrected together with Holm's procedure~\citep{holm1979simple}: over
the whole pool for $W$ and $\mathrm{WC}$, and over the optimizer finals for $\widehat\Delta_{g_0}$,
unless stated otherwise. A contrast is \emph{significant} when it survives this correction and
\emph{distinguishable} when its interval excludes zero. $B(N)$, $C_{95}(N)$ and the Reality Check
$p$-value come from the same resamples. For each prompt, we subtract its observed contrast from each
bootstrap contrast. We then record the largest of the $N$ centered contrasts in each resample.
Using the same resampled examples for all prompts retains their dependence.

\paragraph{Fixed choices and preregistration.}
Each run fixes how it picks its final and whether it penalizes prompt length
(Appendix~\ref{app:pool}); Appendix~\ref{app:protocol} varies these choices. Five studies after the
CivilComments one were preregistered with their predictions (Appendix~\ref{app:prereg}).

\paragraph{Stability and transfer.}
To measure split-half sign replication, we identify differences whose intervals exclude zero on one
half and check their sign on the other. To isolate the effect of the number of groups, we randomly
repartition the same examples within each label while keeping the predictions and the total test
budget fixed. Appendix~\ref{app:ceiling} details both analyses.
To measure transfer, we run the four prompts with significant gains on the initially worst group
unchanged on GPT-4o-mini and Qwen3-235B. We also directly compare the effects between models on
common valid examples (Appendix~\ref{app:second-model}).

\paragraph{Use of generative AI.}
The task and optimizer models described above were used to generate predictions and candidate
prompts, respectively. OpenAI Codex assisted with reviewing and reorganizing the manuscript and
with its LaTeX preparation.

\subsection{Ethical Considerations}

We use three public datasets, CivilComments, MultiNLI and the multilingual toxicity collection of
TextDetox, under their licenses, and collect no new human data or annotations. Two of them contain
toxic and offensive comments, many of them targeting the identity groups we study. The group labels are those of the dataset annotators, and we use them only to compute accuracy
per group. The paper is about measurement and proposes no prompt for deployment. Its main concern is
a risk on the side of evaluation: a worst-group number that inflates a gain can suggest that a
toxicity classifier serves a group better than it does, and one that deflates a gain can hide an
improvement for the group a method was meant to help. The checks of Section~\ref{sec:checks} are
meant to make both visible. All model queries together cost \$89.82 (Appendix~\ref{app:repro}).

\section{Results}
\label{sec:results}
\subsection{Is the Winner Real?}
\label{sec:winner}

\begin{figure}[!htbp]
  \centering
  \includegraphics[width=\linewidth]{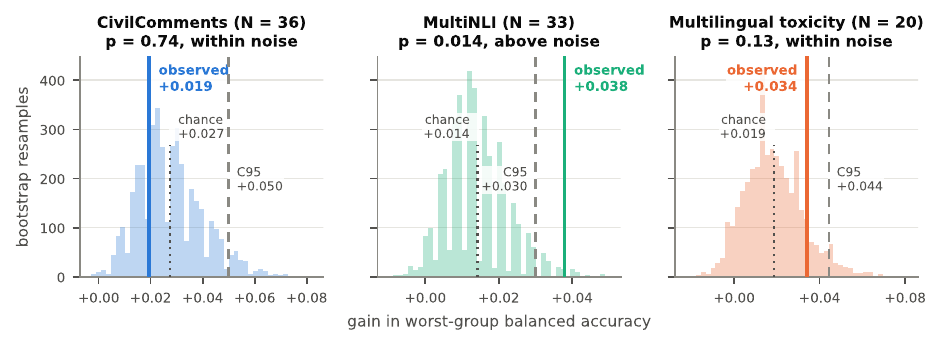}
  \caption{\textbf{The gain of the best prompt against the gain that chance alone gives the best of $N$ prompts.}
    Each histogram shows the best-of-$N$ gain in worst-group balanced accuracy under the null that
    no prompt is better than the seed. The dotted line marks its mean, the chance baseline; the
    solid line marks the observed gain. The dashed line marks the 95th percentile of the null
    distribution, $C_{95}$. The $p$-value is the share of null bootstrap resamples in which the
    best-of-$N$ gain is at least as large as the observed gain (solid line). $N$ is the pool size.
    Histograms use 4,000 paired bootstrap resamples. Each bootstrap contrast is centred by subtracting its observed value
    (Section~\ref{sec:setup}).}
  \label{fig:winners-curse}
\end{figure}

We ask two questions of every pool: whether any single prompt is distinguishable or significant
against the seed (Section~\ref{sec:experiments}), and whether the gain of the best prompt,
$\widehat\Delta_{\max}$, exceeds the chance baseline $B(N)$ and the noise ceiling $C_{95}(N)$
(Table~\ref{tab:winner}, Figure~\ref{fig:winners-curse}). The first concerns each prompt, the second
the number that a comparison of methods reports.

\begin{table}[H]
  \caption{\textbf{The reported gain against the quantities of Section~\ref{sec:formulation}.} Every
    prompt against the seed on the same test examples, Holm correction over the pool. ``Finals only'' repeats the Reality Check
    without the one-line edits. The block without groups counts the contrasts that survive Holm
    correction over the pool, split by sign. The block on $g_0$ covers the optimizer finals and is
    corrected over them. $^\dagger$Nine of the ten are one-line edits.}
  \label{tab:winner}
  \centering
  \small
  \setlength{\tabcolsep}{2.5pt}
  \begin{tabular}{@{}lccc@{}}
    \toprule
     & \begin{tabular}[b]{@{}c@{}}Civil-\\Comments\end{tabular}
     & MultiNLI
     & \begin{tabular}[b]{@{}c@{}}Multilingual\\toxicity\end{tabular} \\
    \midrule
    \multicolumn{4}{@{}l}{\emph{Worst-group balanced accuracy, all prompts}} \\
    Pool size $N$                      & 36 & 33 & 20 \\
    Mean CI width                      & 0.050 & 0.039 & 0.060 \\
    Distinguishable                    & 0 & 13 & 2 \\
    Significant                        & 0 & 2 & 0 \\
    Best gain $\widehat\Delta_{\max}$  & $+0.019$ & $+0.038$ & $+0.034$ \\
    Chance baseline $B(N)$             & $+0.027$ & $+0.014$ & $+0.019$ \\
    Noise ceiling $C_{95}(N)$          & $+0.050$ & $+0.030$ & $+0.044$ \\
    Reality Check $p$                  & 0.74 & 0.014 & 0.13 \\
    \quad finals only                  & 0.74 & 0.013 & 0.12 \\
    \midrule
    \multicolumn{4}{@{}l}{\emph{No groups: significant gains / losses}} \\
    BA on the whole test               & 0 / 5 & 15 / 7 & 0 / 2 \\
    Worst class $\mathrm{WC}$          & 14 / 17 & 19 / 9 & 0 / 10$^\dagger$ \\
    \midrule
    \multicolumn{4}{@{}l}{\emph{Initially worst group $g_0$, optimizer finals}} \\
    Finals                             & 25 & 22 & 9 \\
    Significant $\widehat\Delta_{g_0}$ & 0 & 2 & 2 \\
    Best $\widehat\Delta_{g_0}$        & $+0.030$ & $+0.038$ & $+0.061$ \\
    \bottomrule
  \end{tabular}
\end{table}

Only MultiNLI passes both tests. On CivilComments no prompt of the pool is even distinguishable from
the seed, and the best gain, $+0.019$, falls short of the $+0.027$ that the best of 36 prompts gains
by chance alone; noise reaches it in three resamples out of four. The pool stays within chance when
it is run on GPT-4o-mini as well (Appendix~\ref{app:second-model}). On MultiNLI two prompts are
significant, and the best gain is 2.7 times the chance baseline. Multilingual toxicity stays within
chance, although Section~\ref{sec:moves} shows that a real gain is hidden there. Removing the
one-line edits from the pool changes none of these verdicts (``finals only'' in
Table~\ref{tab:winner}).

\paragraph{Control on the same predictions.}
A null result on CivilComments could mean that the prompts change nothing, or that a gain is lost
when the test is split into groups. We check both possibilities on the same predictions. The prompts do change the
model: worst-class accuracy, computed without groups, separates 31 of the 36 prompts from the seed,
14 upward and 17 downward, as the prompts shift the balance between recall and specificity.
Balanced accuracy counts such a shift only net of its cost to the other class, and by that measure
no prompt significantly improves on the whole test: the best gain, $+0.007$, is below its own chance baseline
of $+0.008$, and the five significant contrasts are losses (Table~\ref{tab:winner}). On
CivilComments neither metric detects a significant gain. The same comparison separates the
other two benchmarks. On MultiNLI balanced accuracy on the whole test finds 15 significant gains,
while the minimum over ten genres finds two (Section~\ref{sec:stable}); on multilingual toxicity it
finds none, while the initially worst group gains significantly (Section~\ref{sec:moves}).
Appendix~\ref{app:ceiling} compares these metrics on all three benchmarks, together with the
worst-group accuracy of WILDS.

\paragraph{Selection inside the optimizer.}
Nor does the optimizer's own choice of final explain the null result. On multilingual toxicity the
best candidate in a run's archive, picked on one half of the test and compared on the other, is
significantly better than the final in none of the 12 runs (Appendix~\ref{app:validation}).

\subsection{When the Minimum Moves}
\label{sec:moves}

The gain on the initially worst group is an upper bound on the reported gain, $\widehat\Delta(S) \le \widehat\Delta_{g_0}(S)$ (Section~\ref{sec:formulation}). The inequality is strict when a prompt lifts $g_0$ above another group: that group becomes the minimum, and the reported gain captures only part of the improvement, or none of it.

\begin{figure}[!t]
  \centering
  \includegraphics[width=0.94\linewidth]{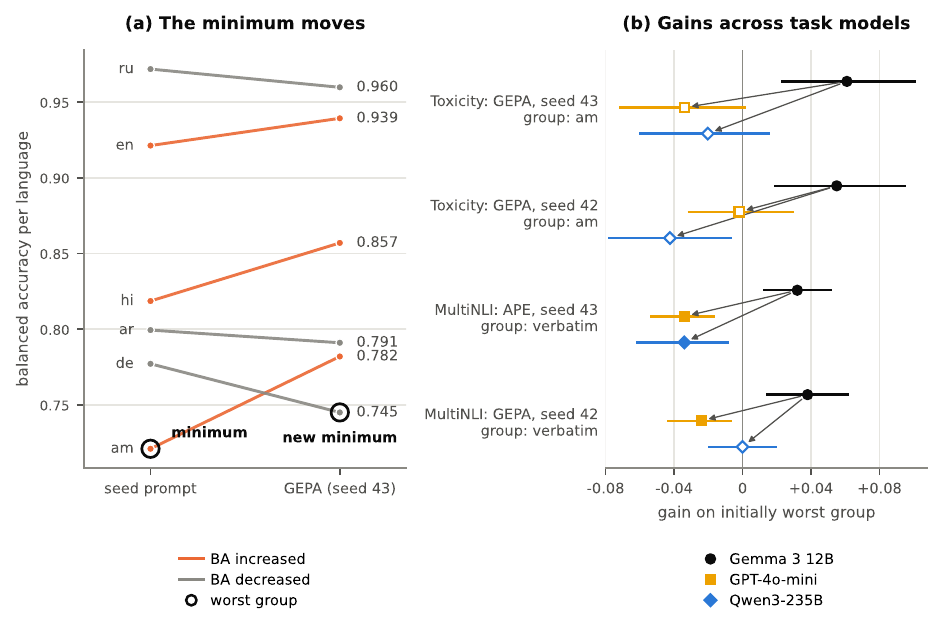}
  \caption{\textbf{A moving minimum hides a gain, and surviving gains do not transfer between task models.}
    (a) Balanced accuracy per language under the seed and the GEPA final of seed 43 on multilingual
    toxicity. Orange lines mark improvements. Amharic (am) gains significantly, but German (de)
    becomes the new minimum, so the worst-group gain stays within noise.
    (b) The four finals with significant gains on the initially worst group under Gemma~3 12B,
    run unchanged on GPT-4o-mini and Qwen3-235B. Bars are 95\% bootstrap intervals; filled markers
    are significant after Holm correction. Three significant losses belong to two distinct
    MultiNLI prompts: APE on both alternative models and GEPA on GPT-4o-mini.}
  \label{fig:gain-diagnostics}
  \label{fig:minimum-moved}
  \label{fig:model-transfer}
\end{figure}

This happens on multilingual toxicity (Figure~\ref{fig:minimum-moved}(a)). The GEPA final of seed 43
raises balanced accuracy on Amharic, the worst language under the seed, by $+0.061$. The gain is
significant and comes mostly from recall, as the model now misses far fewer of the toxic Amharic
comments. German drops within noise and becomes the new minimum, so $\widehat\Delta$ for this final
is only $+0.024$. The GEPA final of seed 42 behaves the same way; at $+0.034$ it is the best prompt
of the pool by $\widehat\Delta$, and still within chance (Section~\ref{sec:winner}). Per-language accuracies and the split
into recall and specificity are given in Appendix~\ref{app:mechanism}.

The gain on $g_0$ shows which of the two errors occurs on each benchmark
(Table~\ref{tab:winner}). On CivilComments no final lifts the initially worst group significantly either, so
the null result of Section~\ref{sec:winner} holds for the group that the search was meant to help.
On MultiNLI the inequality holds with equality: \emph{verbatim} stays the worst genre for every
prompt, and both measures find the same two gains. On multilingual toxicity two finals lift Amharic
significantly while $\widehat\Delta$ finds no significant gain. Choosing $g_0$ on the same test set
makes the seed look slightly worse on it and so inflates $\widehat\Delta_{g_0}$, but by less than a
twentieth of these gains (Table~\ref{tab:stable} in Appendix~\ref{app:ceiling}). In addition, Amharic was named as the worst language
before the data were collected (Appendix~\ref{app:prereg}).

A gain on $g_0$ could also come from a mere shift of the decision threshold, with the
model flagging more Amharic comments as toxic. The one-line edits show what such a shift achieves
(Figure~\ref{fig:strictness} in Appendix~\ref{app:mechanism}). The seed-43 final lies inside the range of false-positive rates
that the edits cover and $+0.051$ above the best of them, so at almost the same rate of false alarms
it catches more of the toxic comments. The seed-42 final lies outside that range, and for it the
comparison is an extrapolation. On MultiNLI most finals also lie above the edits, nearly all inside
their range; on CivilComments only five of the 25 do, four of them outside the range.

\subsection{How Stable Is the Worst Group?}
\label{sec:stable}

The minimum in $W$ is itself a selection, and it adds noise that averaged statistics do not have.
Each group is measured on only part of the test, so for the same pool the chance baseline of $W$ is
2.9 to 3.5 times that of balanced accuracy on the whole test (Table~\ref{tab:metrics-cc} in
Appendix~\ref{app:ceiling}). On top of this, the data decide which group attains the minimum, and
the benchmark decides how many groups $K$ there are. This section measures how each of the two
choices affects the chance baseline and replication.

\paragraph{Which group is the minimum.}
On CivilComments the observed worst group remains the worst in fewer than half of the bootstrap
resamples, and four of the eight identities are the minimum for some prompt of the pool. On MultiNLI
and multilingual toxicity it remains the worst in about nine resamples out of ten
(Table~\ref{tab:stable} in Appendix~\ref{app:ceiling}). An unstable minimum raises the chance
baseline. Intervals on CivilComments are only about a quarter wider than on MultiNLI
(Table~\ref{tab:winner}), yet for the same number of prompts $B(N)$ on CivilComments is about twice
as high (Figure~\ref{fig:ceiling-vs-n} in Appendix~\ref{app:ceiling}). When the minimum moves between groups, different prompts
are compared on different groups, their contrasts are less correlated across resamples, and the
largest of $N$ of them grows faster with $N$. A minimum over noisy estimates also falls below the
true minimum on average, most of all where it is unstable (Table~\ref{tab:stable}).

\paragraph{Replication.}
The minimum replicates least on
all three benchmarks. On CivilComments only 60\% of the distinguishable differences it finds hold on the other half, against 89\% for mean balanced accuracy. It also finds the fewest distinguishable differences, and none of
them is distinguishable with the opposite sign on the other half. What changes between halves is
therefore which differences become distinguishable, while their sign holds. The smoothed statistics
replicate better because they rank prompts almost as the mean does, so smoothing the minimum is no
remedy (Appendix~\ref{app:ceiling}). On GPT-4o-mini the minimum replicates even less often on CivilComments
(Appendix~\ref{app:second-model}).

\begin{figure}[!htbp]
  \centering
  \includegraphics[width=0.70\linewidth]{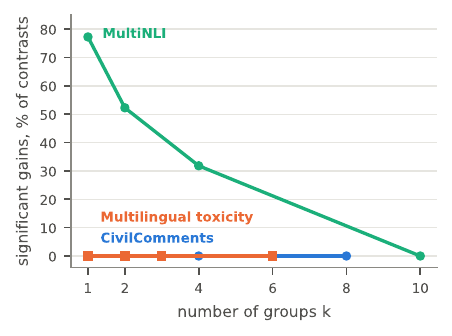}
  \caption{\textbf{More groups leave fewer detectable gains at a fixed test budget.} Fixed examples
    and predictions, randomly partitioned into $k$ equal-sized groups within each class;
    $k=1$ is the whole test. Points show median gains significant after Holm correction, as a
    percentage of 22 MultiNLI finals, 36 CivilComments prompts or 20 toxicity prompts.}
  \label{fig:group-count}
\end{figure}

\paragraph{How many groups.}
The number of groups $K$ changes the result while the effect stays the same. To isolate it, we keep
the test examples and the predictions, split the examples into $k$ random groups of equal size
within each label, and repeat the comparison of Section~\ref{sec:winner}
(Figure~\ref{fig:group-count}); $k = 1$ is the whole test.

On MultiNLI, as $k$ grows from 1 to 10,
the mean size of the contrasts does not change, the intervals widen roughly as $\sqrt{k}$ (Figure~\ref{fig:group-count-width} in Appendix~\ref{app:ceiling}), and the
number of significant gains among the 22 finals falls from 17 to none. At a fixed test budget, every
added group thus costs resolution. On CivilComments and multilingual toxicity there is no
significant gain at any $k$, the whole test included, so the number of groups changes the resolution
of the test without creating a signal where there is none. Groupings with a meaning give the same
verdicts, and coarser groups are no general remedy (Appendix~\ref{app:protocol}).

\subsection{Does the Gain Survive a Change of Model?}
\label{sec:model}

Four finals have a significant gain on the initially worst group $g_0$ under our main task model $M$, Gemma~3 12B: the two GEPA finals on multilingual toxicity, for Amharic, and an APE and a GEPA final on MultiNLI, for \emph{verbatim}. We ran these prompts on the two models $M'$ of Section~\ref{sec:experiments}, GPT-4o-mini and Qwen3-235B, and compared $\widehat\Delta_{g_0}(S; M')$ with $\widehat\Delta_{g_0}(S; M)$ (Figure~\ref{fig:model-transfer}(b); Table~\ref{tab:transfer} in Appendix~\ref{app:second-model}).

Both models are comparable to $M$. Under the seed prompt both are about as accurate as $M$ against the same human labels, and both largely agree with it on which groups are hard. On toxicity and MultiNLI the answers of GPT-4o-mini differ from those of $M$ most on the initially worst group itself (Appendix~\ref{app:second-model}).

None of the four gains holds on either model. On GPT-4o-mini all four change sign, and both MultiNLI gains become significant losses on the same genre. The decision rule fixed in advance for the MultiNLI check, stated in $W$, gives the same verdict. On Qwen3-235B both toxicity gains turn negative, the APE gain on \emph{verbatim} again becomes a significant loss, and the GEPA gain on \emph{verbatim} vanishes. The difference between the models is itself significant for every final: on the examples that all three models answered, the gain on $M$ exceeds the effect on $M'$ by 0.038 to 0.098, and all eight differences survive Holm correction (Table~\ref{tab:transfer}). The threshold argument of Section~\ref{sec:moves} reverses as well: on both models the GEPA finals no longer rise above what the one-line edits and the seed reach on Amharic. Selection on $M$ would pull a gain back toward zero on an independent measurement, but it would not turn the gain into a significant loss. The reversals point to prompts fitted to the error pattern of the main model.

On CivilComments no final has a significant gain to transfer (Section~\ref{sec:moves}), but the
effects of all 25 finals can be compared between $M$ and GPT-4o-mini. Their effects on the initially
worst group are too noisy for this: on GPT-4o-mini they do not replicate across two halves of the
test. Their effects on worst-class accuracy are measured almost without noise, with split-half
reliabilities of 0.99 and 0.96, and still correlate at only $\rho = 0.28$
(Appendix~\ref{app:second-model}). The same prompts shift the balance between the classes
differently on the two models.

\section{Discussion}
\label{sec:checks}

Raising the lowest accuracy across groups and helping the group that started out worst
can favour different prompts. On multilingual toxicity, the GEPA finals from seeds 42 and 43
have estimated gains of 3.4 and 2.4 percentage points in the minimum. On Amharic, the initially
worst language, the respective gains are 5.5 and 6.1 points, reversing the ranking of the point
estimates (Section~\ref{sec:moves}). Reporting both gains shows whether a prompt lifts the
performance floor, helps the initially worst group, or does both.

Table~\ref{tab:checks} gives seven checks for interpreting a reported gain. The first six use
existing predictions. The seventh needs a second task model; it cost less than \$2.5 per benchmark
for the relevant prompts (Appendix~\ref{app:repro}).

\begin{table}[!htbp]
  \caption{\textbf{Seven checks for a worst-group gain from prompt optimization.} Checks 1--6 run on
    predictions already at hand; check 7 needs new queries to a second task model.}
  \label{tab:checks}
  \centering
  \small
  \renewcommand{\arraystretch}{1.12}
  \setlength{\tabcolsep}{3pt}
  \begin{tabular}{@{}c >{\raggedright\arraybackslash}p{0.17\linewidth} >{\raggedright\arraybackslash}p{0.36\linewidth} >{\raggedright\arraybackslash}p{0.32\linewidth} c@{}}
    \toprule
    & Check & Question & What to report & Section \\
    \midrule
    1 & Metrics without groups
      & Do the prompts change the model, and do they raise accuracy before any grouping?
      & $\mathrm{WC}$ and balanced accuracy on the whole test for the same contrasts & \ref{sec:winner} \\
    2 & Random groups
      & Do random groups give the same result as the real grouping?
      & the same comparison on random groups & \ref{sec:stable} \\
    3 & Selection baseline
      & Is the best gain larger than the chance baseline $B(N)$ for $N$ prompts?
      & $N$, $B(N)$ and $C_{95}(N)$ next to $\widehat\Delta_{\max}$ & \ref{sec:winner} \\
    4 & Group stability
      & Does the observed worst group stay the worst under resampling?
      & how often $\arg\min_g \widehat{\mathrm{BA}}_g$ stays the same & \ref{sec:stable} \\
    5 & Number of groups
      & Does the finding hold for another number of groups?
      & $K$ and the interval width at that $K$ & \ref{sec:stable} \\
    6 & Initially worst group
      & Does the minimum hide a gain on the initially worst group?
      & $\widehat\Delta_{g_0}$ next to $\widehat\Delta$ & \ref{sec:moves} \\
    7 & Model transfer
      & Does the gain belong to the task or to the model?
      & $\widehat\Delta_{g_0}(S; M')$ and its difference from $M$; split-half reliability before any correlation & \ref{sec:model} \\
    \bottomrule
  \end{tabular}
\end{table}

Changing the rule for choosing a final can change the reported result even with the same
candidates. On MultiNLI, adding a length penalty cut the mean share of validation draws yielding
a significant test win from 75\% to zero across protocols (Appendix~\ref{app:protocol}). Changing
the selection metric in the archives of completed runs changed the winner in about half the
CivilComments and MultiNLI runs, with small mean changes in test CVaR@25\%
(Appendix~\ref{app:live}). Changing the search objective can also change which candidates are
found. Report the search objective, the rule for choosing the final (including penalties),
and the test metric separately. Method rankings varied across benchmarks.

The noise ceiling of check 3 has to be estimated on the finals themselves. They disagree with the
seed about twice as often as one-line edits do and so have wider intervals, and a ceiling calibrated
on edits is too low on all three benchmarks (Appendix~\ref{app:predict}).

Of 18 preregistered predictions, five held, nine failed and four held in part
(Appendix~\ref{app:prereg}).

\subsection{Limitations}

The model dependence of Section~\ref{sec:model} rests on four gains on two benchmarks, tested on two
further models, and on the effects of the 25 CivilComments finals on one further model; more models
and tasks would show how far it generalizes. On CivilComments there was no significant gain to transfer. All
three benchmarks are binary
classification tasks with balanced test sets, and multilingual toxicity has only nine optimizer
finals. On CivilComments the null result of Section~\ref{sec:winner} and the instability of
Section~\ref{sec:stable} hold on GPT-4o-mini as well; the other results of Sections~\ref{sec:winner}
to~\ref{sec:stable} describe the main task model.
The instability of the worst group under resampling is pronounced on CivilComments only. Budgets on
CivilComments differ across seeds for four methods and PRIME used more labelled data than the other
methods, so the paper makes no claim about which method is best. The WILDS metric is computed on our
assignment of comments to identities rather than on the overlapping groups of WILDS. The test of
selection inside the optimizer covers one benchmark and one task model and rules out only regrets
larger than 0.011. The protocol analysis simulates selection over a fixed pool, while the live
searches cover three methods per benchmark. Finally, hosted models change over time, so the exact
numbers may not be reproducible with later versions served under the same model identifiers; the
saved predictions allow every analysis in the paper to be rerun without new queries.

\section{Conclusions}
\label{sec:conclusion}

A worst-group gain combines changes in group performance with two selections: the best prompt in
the pool and the lowest-scoring group. On CivilComments the selected gain was within the range of
chance; on multilingual toxicity a moving minimum masked an improvement on Amharic; and none
of the four group-specific gains transferred to either alternative task model. Reporting the selection
baseline, the gain on the initially worst group, and the effect on another task model distinguishes
these cases. When one scalar summarizes several objectives, report the selections that produced it.

\paragraph{Author Contributions.} Software, investigation, and writing---original draft preparation, A.M.; supervision, O.S.; writing---review and editing, A.M. and O.S.
\paragraph{Funding.} This work was supported by AIRI and the Moscow Institute of Physics and Technology (MIPT).
\paragraph{Data Availability Statement.} The source datasets are available from Hugging Face as \href{https://huggingface.co/datasets/pietrolesci/civilcomments-wilds}{CivilComments (WILDS)}, \href{https://huggingface.co/datasets/nyu-mll/multi_nli}{MultiNLI}, and \href{https://huggingface.co/datasets/textdetox/multilingual_toxicity_dataset}{TextDetox multilingual toxicity}. The \href{https://github.com/marcheanin/worst-group-gain/tree/41e077059772e85587bfe191ed7500a5d324bd53}{public research repository at commit \texttt{41e0770}} contains analysis code, the split indices, labels, group identifiers, saved predictions, and the preregistration plans. Appendix~\ref{app:repro} describes which analyses the archived code covers.
\paragraph{Conflicts of Interest.} The authors declare no conflicts of interest. The funders had no role in the study design, data collection, analysis or interpretation, manuscript preparation, or the decision to publish.

\appendix
\section{Preregistrations}
\label{app:prereg}

After the CivilComments study we ran five follow-up studies, and for each we wrote down, before
collecting its data, what we expected and the criterion that would decide it: MultiNLI as a second
benchmark, the mechanism behind the CivilComments result, multilingual toxicity as a controlled copy
of CivilComments, selection inside the optimizer (Appendix~\ref{app:validation}), and the transfer
of gains to a second task model (Section~\ref{sec:model}). The dated plans are in the public
repository linked in the Data Availability Statement.
Table~\ref{tab:prereg} lists all 18 predictions, grouped by the part of the paper they
bear on.

Five predictions held, nine failed and four held in part. The predictions were written after
extensive work with these measurements and still failed more often than they held, which is one
reason this paper recommends checks that run on the predictions themselves. Several results of the
main text rest on criteria fixed before the data. On MultiNLI, findings of the worst-group metric
replicate worse than those of smoothed statistics; on CivilComments, random groups create no signal;
on MultiNLI the worst group is the same for every final, and on multilingual toxicity it changes for
only three of nine (Section~\ref{sec:stable}). The optimizer's own selection does not
lose better candidates (Section~\ref{sec:winner}). On MultiNLI, neither of the two significant gains
transfers to the second model, the outcome of the decision rule fixed in that plan
(Section~\ref{sec:model}).

Every prediction is scored by the criterion written before the data. For the main toxicity
prediction this matters: a threshold recomputed after the outcome would have turned the miss into a
hit, and we count it as a miss. Two analyses depart from their plans in implementation only. One
first used nine random groups instead of eight and was rerun with eight, with the same result; the
handling of unparsable answers in the selection study is described in Appendix~\ref{app:validation}.

\begin{table}[H]
  \caption{\textbf{The 18 preregistered predictions}, grouped by the part of the paper they bear on.
    BA is balanced accuracy; CVaR is the mean over the worst quarter of groups. A prediction held if
    every part of its criterion was met, failed if the refutation criterion of its plan was met, and
    held in part otherwise. $^\ddagger$The planned threshold
    curve included the finals, so the gain it measured equalled that of the best final by
    construction. On the curve of the one-line edits alone (Section~\ref{sec:moves}) it is $+0.012$.}
  \label{tab:prereg}
  \centering
  \footnotesize
  \setlength{\tabcolsep}{4pt}
  \begin{tabularx}{\linewidth}{@{}>{\raggedright\arraybackslash}p{6.0cm} >{\raggedright\arraybackslash}X l@{}}
    \toprule
    Prediction & Criterion fixed in advance $\rightarrow$ observed & Verdict \\
    \midrule
    \multicolumn{3}{@{}l}{\emph{Is the winner real? (Section~\ref{sec:winner})}} \\
    MultiNLI is almost as unresolvable as CivilComments
      & $\leq$10\% of contrasts significant after Holm, by worst-group BA and by CVaR $\rightarrow$ 6.1\% and 30.3\%
      & partial \\
    Multilingual toxicity is unresolvable
      & best gain $\leq +0.031$, the noise ceiling set before the optimizer runs $\rightarrow$ $+0.034$
      & failed \\
    On multilingual toxicity the prompts change the model: worst-class accuracy separates the finals
      & $\geq$6 of 12 finals $\rightarrow$ 1 of 9
      & failed \\
    The toxicity seed prompt is weak, with Amharic as its worst group
      & worst-group BA in [0.55, 0.78], worst group Amharic $\rightarrow$ 0.721, Amharic
      & held \\
    The optimizer's selection loses a better candidate in some run
      & significant after Holm in $\geq$1 of 12 runs $\rightarrow$ 0 of 12
      & failed \\
    That loss matters on average
      & cross-fit regret $\geq$0.02 $\rightarrow$ $-0.001$
      & failed \\
    \midrule
    \multicolumn{3}{@{}l}{\emph{Threshold shift or better discrimination (Section~\ref{sec:moves})}} \\
    The MultiNLI gain lifts the curve rather than shifting the threshold
      & gain of the best threshold shift over the seed $<$0.015 $\rightarrow$ $+0.038$$^\ddagger$
      & failed \\
    \midrule
    \multicolumn{3}{@{}l}{\emph{How stable is the worst group? (Section~\ref{sec:stable})}} \\
    Worst-group statistics replicate worse than smoothed ones (MultiNLI)
      & gap $\geq$5 points $\rightarrow$ 82.0\% vs.\ 97.2\%
      & held \\
    On MultiNLI the worst group changes between finals
      & it changes for $\geq$8 of 22 finals $\rightarrow$ 0 of 22
      & failed \\
    On multilingual toxicity the worst group is stable
      & it changes for $\leq$3 finals, $\leq$2 languages are ever the minimum $\rightarrow$ 3 of 9, 2 languages
      & held \\
    On multilingual toxicity the two worst groups are far apart
      & gap $\geq$0.030 $\rightarrow$ 0.056
      & held \\
    On CivilComments random groups create no signal
      & median Holm count over 200 random partitions $\leq$1 $\rightarrow$ 0
      & held \\
    On MultiNLI random groups raise resolvability
      & median $\geq$5 of 22 significant $\rightarrow$ 0
      & failed \\
    Regrouping MultiNLI by difficulty gives no gain
      & $\leq$1 significant and $|\Delta| <$ 0.015 $\rightarrow$ 5 and 10 significant, $|\Delta|$ 0.011 and 0.017
      & partial \\
    Larger cells give narrower intervals and no fewer findings
      & intervals narrow as cells grow; findings do not drop $\rightarrow$ intervals as predicted; findings 0, 2, 0
      & partial \\
    A new draw of the test changes the number of findings at constant width
      & ratio of finding shares $\geq$3 at a ratio of widths $\leq$1.15 $\rightarrow$ 2.14 at 1.04
      & partial \\
    \midrule
    \multicolumn{3}{@{}l}{\emph{Protocol and predictability (Section~\ref{sec:checks}, Appendices~\ref{app:protocol} and~\ref{app:predict})}} \\
    Selection protocols yield many point wins but no significant ones
      & range of point wins $\geq$30 points, none significant $\rightarrow$ range 5 points, median 97\% significant (CVaR)
      & failed \\
    The leverage of the threshold on toxicity is as on CivilComments
      & leverage in [4, 7] $\rightarrow$ 2.51
      & failed \\
    \bottomrule
  \end{tabularx}
\end{table}

\section{Benchmarks and Task Models}
\label{app:benchmarks}

\paragraph{Data.}
CivilComments is the WILDS release of the dataset (\href{https://huggingface.co/datasets/pietrolesci/civilcomments-wilds}{Hugging Face},
test split). A comment belongs to the first of the eight identities annotated with a score of at
least 0.5; the test set also holds 400 comments, sampled as a ninth stratum, that mention none of
them. These enter worst-class and overall accuracy but no group metric. MultiNLI uses the matched
and mismatched validation sets (\href{https://huggingface.co/datasets/nyu-mll/multi_nli}{Hugging Face}); the positive class is contradiction,
the negative class entailment and neutral, and the group is the genre field. The training split of
MultiNLI covers only the five matched genres, so the optimizers draw no training examples from the
other five, including \emph{verbatim}; the validation data on which they score candidates cover all
ten genres. Multilingual toxicity takes six language splits of the
\href{https://huggingface.co/datasets/textdetox/multilingual_toxicity_dataset}{TextDetox dataset}. Every test set is drawn with the same number of
examples per group--label cell: 200, 250 and 250. The CivilComments seed prompt and its eleven
one-line edits are reused on multilingual toxicity byte for byte (all twelve SHA-256 hashes match);
MultiNLI has its own seed and its own eleven edits, built the same way. CivilComments is released under
CC0, MultiNLI under the open licenses listed by its authors, and the multilingual toxicity collection
under OpenRAIL++.

\paragraph{Task models.}
We chose the main task model in a pilot on CivilComments because its errors under the seed prompt
were spread over both classes, which leaves a prompt room to move it in either direction.
All task models are queried with temperature 0 and at most 16 output tokens
(Table~\ref{tab:task-models}). The answer is parsed by a chain of regular expressions: an explicit
``label: 0/1'', then the label words, then a bare digit. An answer that matches none of them is
marked unparsable rather than read as the negative class, and the example is then dropped for every
prompt of the benchmark.

\begin{table}[!htbp]
  \caption{\textbf{Task models}, all queried through OpenRouter with temperature 0, at most 16 output
    tokens and the same fail-closed parser; the providers of the replication models are pinned.
    Test predictions were collected in September 2026 for Gemma~3 12B and GPT-4o-mini and in October
    2026 for both Qwen3 models; \texttt{qwen3-235b-a22b-2507} and \texttt{qwen3-30b-a3b-2507} are the
    Qwen3-235B-A22B-Instruct-2507 and Qwen3-30B-A3B-Instruct-2507 releases.
    $^\ddagger$MultiNLI only; on toxicity it failed the admission rule (Appendix~\ref{app:second-model}).
    $^\dagger$The only provider of the model on OpenRouter; it
    is logged for the MultiNLI and toxicity runs, but the logger of the CivilComments run did not
    record it.}
  \label{tab:task-models}
  \centering
  \footnotesize
  \setlength{\tabcolsep}{3pt}
  \begin{tabular}{@{}lll@{}}
    \toprule
    Model & Role & Provider \\
    \midrule
    \texttt{gemma-3-12b-it}       & main        & DeepInfra, bf16$^\dagger$ \\
    \texttt{gpt-4o-mini}          & replication & OpenAI \\
    \texttt{qwen3-235b-a22b-2507} & replication & DeepInfra, fp8 \\
    \texttt{qwen3-30b-a3b-2507}   & replication$^\ddagger$ & Alibaba \\
    \bottomrule
  \end{tabular}
\end{table}

\paragraph{Unparsable answers.}
Table~\ref{tab:lost-rows} shows how many test examples were dropped and why. Every dropped pair of
prompt and example was queried again. On CivilComments and MultiNLI almost all failures were
transport errors that parse on a retry. On multilingual toxicity, one EvoPrompt-DE final gives no
label on 79 English examples, 16\% of the English test set, and on 73 of them it repeats the markup of
its own prompt on a retry as well. The paper reports the preregistered analysis, which drops these examples.
Recovering the transport failures, or keeping all toxicity examples and scoring the broken final's
answers as wrong, changes no verdict; the largest shift is in the toxicity Reality Check $p$, from
0.134 to 0.172.

\begin{table}[!htbp]
  \caption{\textbf{Test examples dropped because of unparsable answers, and the verdicts with and
    without them.} ``Transport'': the answer parsed on a retry. ``Broken'': the prompt itself
    produces no label. ``Recovered'': transport failures replaced by the answers of the retry. ``All
    kept'': the 3,000 toxicity examples, with the broken final's unparsable answers scored as wrong.
    The noise ceiling moves by at most 0.002.}
  \label{tab:lost-rows}
  \centering
  \footnotesize
  \setlength{\tabcolsep}{3pt}
  \begin{tabular}{@{}lcccc ccc@{}}
    \toprule
    & & & & & \multicolumn{3}{c}{Holm count; best gain; Reality Check $p$} \\
    \cmidrule(l){6-8}
    Benchmark & Dropped & \shortstack{Pairs\\re-queried} & Transport & Broken & as reported & recovered & all kept \\
    \midrule
    CivilComments & 16 of 3,600 & 16 & 15 & 1  & 0; $+0.019$; 0.743 & 0; $+0.020$; 0.725 & --- \\
    MultiNLI      & 9 of 5,000  & 9  & 9  & 0  & 2; $+0.038$; 0.014 & 2; $+0.038$; 0.010 & --- \\
    Toxicity      & 93 of 3,000 & 94 & 21 & 73 & 0; $+0.034$; 0.134 & 0; $+0.032$; 0.172 & 0; $+0.032$; 0.169 \\
    \bottomrule
  \end{tabular}
\end{table}

\section{Prompt Pool and Budgets}
\label{app:pool}

\paragraph{Where the finals come from.}
On CivilComments the finals come from a matrix of separate runs: eleven method variants, each with
up to three seeds (Table~\ref{tab:pool-cc}). The baseline runs pick their finals by a smoothed
worst-group balanced accuracy on their own validation data. A run whose final is byte-identical to the seed, or to
a prompt already in the pool, adds nothing. On MultiNLI and multilingual toxicity the finals come
from live optimization loops (Table~\ref{tab:pool-live}). The pool also holds the eleven one-line
edits; their manifest has a twelfth entry, a deliberate copy of the seed kept as a calibration
anchor, which is not counted.

\paragraph{One-line edits.}
The seed prompt ends with a rule for unsure cases. On CivilComments and multilingual toxicity it
reads ``When unsure, prefer 0 (non-toxic) unless clear abusive intent'', and on MultiNLI ``When
unsure, prefer 0 (not a contradiction) unless the incompatibility is clear''. Each one-line edit
replaces this rule and leaves the rest of the prompt unchanged (Table~\ref{tab:edits}). The MultiNLI
edits follow the same pattern for contradiction, for instance ``Label 1 only if the probability that
the hypothesis contradicts the premise is at least 90\%; otherwise label 0''.

\begin{table}[!htbp]
  \caption{\textbf{The eleven one-line edits} on CivilComments and multilingual toxicity, where they
    are identical. Only this rule of the seed prompt is changed.}
  \label{tab:edits}
  \centering
  \footnotesize
  \setlength{\tabcolsep}{3pt}
  \begin{tabular}{@{}>{\raggedright\arraybackslash}p{1.9cm}>{\raggedright\arraybackslash}p{5.5cm}@{}}
    \toprule
    Edit & Rule that replaces the seed's rule for unsure cases \\
    \midrule
    Threshold $X$ (8 edits) & Label 1 only if the probability that the comment is toxic is at least
      $X$\%; otherwise label 0 \quad ($X = 20, 30, \ldots, 90$) \\
    Most lenient & When unsure, label 0. Reserve label 1 for comments that are unambiguously
      abusive \\
    Strict & When unsure, prefer 1 (toxic); err on the side of flagging \\
    Strictest & When unsure, label 1. Flag any comment that a member of the targeted group could
      reasonably find demeaning \\
    \bottomrule
  \end{tabular}
\end{table}

\begin{table}[H]
  \caption{\textbf{CivilComments finals by seed.} ``new'': a final that differs from the seed;
    ``seed'': the run returned the seed; ``dup.'': identical to the ProTeGi final of the same
    seed; ``---'': not run.}
  \label{tab:pool-cc}
  \centering
  \small
  \setlength{\tabcolsep}{3pt}
  \begin{tabular}{@{}lcccc@{}}
    \toprule
    Variant & 42 & 43 & 44 & Finals \\
    \midrule
    APE                          & new & seed & new & 2 \\
    APE, 48 proposals            & new & new  & new & 3 \\
    APE, unlabelled target demos & seed & new & new & 2 \\
    ProTeGi                      & new & new  & new & 3 \\
    ProTeGi, all labels          & new & ---  & --- & 1 \\
    ProTeGi, random labels       & new & dup. & new & 2 \\
    GPO                          & new & new  & new & 3 \\
    EvoPrompt-GA                 & new & new  & new & 3 \\
    EvoPrompt-DE                 & new & new  & new & 3 \\
    GEPA                         & new & seed & seed & 1 \\
    PRIME                        & new & seed & new & 2 \\
    \midrule
    Total                        & & & & 25 \\
    \bottomrule
  \end{tabular}
\end{table}

\begin{table}[H]
  \caption{\textbf{Live optimization loops.} On MultiNLI, five runs (all GEPA) returned the seed;
    the 22 finals are 9 from APE, 9 from EvoPrompt-DE and 4 from GEPA. On toxicity, three runs
    returned the seed, one each of EvoPrompt-DE, APE and GEPA. The selection statistics on MultiNLI
    are worst-group balanced accuracy, its smoothed version and overall accuracy; on toxicity runs
    select by worst-group balanced accuracy.}
  \label{tab:pool-live}
  \centering
  \small
  \setlength{\tabcolsep}{3pt}
  \begin{tabular}{@{}>{\raggedright\arraybackslash}p{2.0cm}>{\raggedright\arraybackslash}p{3.1cm}cc@{}}
    \toprule
    Benchmark & Design & Runs & Finals \\
    \midrule
    MultiNLI & APE, EvoPrompt-DE, GEPA $\times$ 3 selection statistics $\times$ 3 seeds & 27 & 22 \\
    Toxicity & APE, ProTeGi, EvoPrompt-DE, GEPA $\times$ 3 seeds & 12 & 9 \\
    \bottomrule
  \end{tabular}
\end{table}

\paragraph{Implementations.}
All methods run through one interface: the same optimizer LLM (DeepSeek-V4-Pro) proposes and edits
prompts, the same task model scores them, and each run picks its final by the statistic fixed in its
configuration. ProTeGi and GPO are ported from the authors' code, and APE follows its implementation
in the GPO code; EvoPrompt and GEPA are reimplemented from the papers. For GEPA we implement its core
loop, reflective mutations from error feedback with a Pareto archive over per-example correctness,
capped by the number of metric calls, rather than the authors' package.

\paragraph{Budgets.}
Table~\ref{tab:budgets} gives the budget of every method, read from the saved trace of each run.

\begin{table}[H]
  \caption{\textbf{Budgets of the optimizers}, read from the trace of each run.}
  \label{tab:budgets}
  \centering
  \small
  \setlength{\tabcolsep}{3pt}
  \begin{tabular}{@{}>{\raggedright\arraybackslash}p{3.0cm}>{\raggedright\arraybackslash}p{3.6cm}>{\raggedright\arraybackslash}p{4.6cm}>{\raggedright\arraybackslash}p{1.9cm}>{\raggedright\arraybackslash}p{1.9cm}@{}}
    \toprule
    Method & Budget & CivilComments & MultiNLI & Toxicity \\
    \midrule
    APE & proposals, demonstrations & 6, 36 & 6, 36 & 6, 36 \\
    APE, 48 proposals & proposals, demonstrations & 48, 288 & --- & --- \\
    APE, unlabelled target demos & proposals, demonstrations, unlabelled & 6, 36, 12 & --- & --- \\
    ProTeGi & rounds $\times$ beam, minibatch & $6\times4$ (seed 42), $2\times4$ (43, 44); 64 & --- & $2\times4$; 64 \\
    ProTeGi, all labels & rounds $\times$ beam, minibatch & $2\times4$; 64 & --- & --- \\
    ProTeGi, random labels & rounds $\times$ beam, minibatch & $4\times4$ (42), $2\times4$ (43, 44); 64 & --- & --- \\
    GPO & proposals, demonstrations, confidence threshold & 6, 36, 0.83 & --- & --- \\
    EvoPrompt-GA & population $\times$ generations, validation examples & $6\times5$ (42), $4\times3$ (43, 44); 500 & --- & --- \\
    EvoPrompt-DE & population $\times$ generations, validation examples & $4\times3$; 500 & $4\times3$; 500 & $4\times3$; 500 \\
    GEPA & reflections, metric-call limit, validation examples & 12 (42), 8 (43, 44); 20,000; 400 & 8; 20,000; 400 & 8; 20,000; 400 \\
    PRIME & labelled examples & 972 training, 900 validation & --- & --- \\
    \bottomrule
  \end{tabular}
\end{table}

 On
CivilComments the baselines draw on 360 labelled training examples from comments that mention no
identity or the male, female or christian identities, plus 4,000 unlabelled examples from the LGBTQ,
muslim, other-religion, black and white identities; the variant with random labels
may label 240 of them, and the diagnostic ProTeGi run sees all labels. On MultiNLI the baselines see
360 labelled examples from the five training genres, and on toxicity 432. PRIME used 972 labelled
training examples and 900 labelled validation examples (Appendix~\ref{app:prime}).

Two things follow. First, within the CivilComments matrix the budget is not the same across seeds for
ProTeGi, ProTeGi with random labels, EvoPrompt-GA and GEPA: seed 42 got the larger one. Both GEPA
runs with the smaller budget returned the seed. We make no claim about which method is better on
CivilComments, so this does not touch our conclusions. Second, the live loops of
Appendix~\ref{app:live}, on which the ranking of methods across benchmarks rests
(Section~\ref{sec:checks}), give APE, EvoPrompt-DE and GEPA the same budgets on all three benchmarks,
so a different behaviour of a method on different benchmarks is not a matter of budget. On toxicity the budget of EvoPrompt-DE is equal only nominally: half of its archive slots
hold candidates whose answers cannot be parsed (Appendix~\ref{app:validation}).

\section{Chance Baseline, Argmin and Replicability}
\label{app:ceiling}

\begin{figure}[H]
  \centering
  \includegraphics[width=0.82\linewidth]{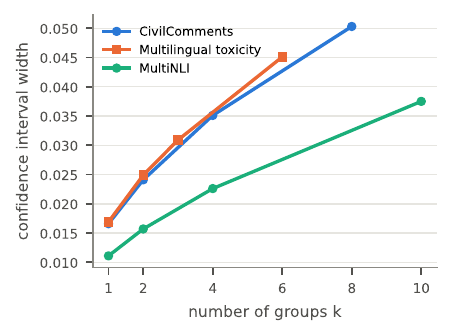}
  \caption{\textbf{Finer random partitions widen uncertainty at a fixed test budget.} The same
    examples and predictions as in Figure~\ref{fig:group-count}, regrouped within each class.
    Points show the median width of the 95\% confidence intervals over random partitions;
    $k=1$ is the whole test. Interval width grows roughly as $\sqrt{k}$.}
  \label{fig:group-count-width}
\end{figure}

\paragraph{Metrics on the same predictions.}
Table~\ref{tab:metrics-cc} applies the analysis of Section~\ref{sec:winner} to two metrics without
groups computed on the same predictions, next to worst-group balanced accuracy. Balanced accuracy on
the whole test is computed as accuracy over all test examples; every test set holds equal numbers of
each label, so up to the few dropped examples the two coincide. On CivilComments worst-class
accuracy, a minimum over the two classes, separates 31 of the 36 contrasts after Holm correction, 14
of them gains, and its best gain is 3.8 times its own chance baseline. Balanced accuracy on the whole
test separates five, all of them losses, and its best gain is no larger than chance. Worst-group
balanced accuracy separates none. The prompts therefore shift the balance between recall and
specificity without raising balanced accuracy, so the worst-group metric has no gain to detect even
before the test is split into groups; after the split it no longer detects the five losses either.
On MultiNLI balanced accuracy on the whole test finds 15 significant gains and the worst-group metric
two, so there the split into ten genres costs most of the findings. On multilingual toxicity neither
finds a gain. On all three benchmarks the chance baseline of the worst-group metric is 2.9 to 3.5
times that of balanced accuracy on the whole test.

\begin{table}[!htbp]
  \caption{\textbf{Metrics with and without groups on the same predictions}, all prompts of the pool
    against the seed. Half-width of the 95\% interval, best gain, chance baseline, contrasts
    significant after Holm correction (gains / losses) and Reality Check $p$.}
  \label{tab:metrics-cc}
  \centering
  \small
  \setlength{\tabcolsep}{2.5pt}
  \begin{tabular}{@{}lccccc@{}}
    \toprule
    Metric & \begin{tabular}[b]{@{}c@{}}Half-\\width\end{tabular} & \begin{tabular}[b]{@{}c@{}}Best\\gain\end{tabular}
      & \begin{tabular}[b]{@{}c@{}}Chance\\baseline\end{tabular} & Holm & $p$ \\
    \midrule
    \multicolumn{6}{@{}l}{\emph{CivilComments, 36 contrasts}} \\
    Worst class     & 0.014 & $+0.055$ & $+0.014$ & 14 / 17 & 0.001 \\
    BA, whole test  & 0.008 & $+0.007$ & $+0.008$ & 0 / 5   & 0.534 \\
    Worst-group BA  & 0.025 & $+0.019$ & $+0.027$ & 0 / 0   & 0.743 \\
    \midrule
    \multicolumn{6}{@{}l}{\emph{MultiNLI, 33 contrasts}} \\
    Worst class     & 0.009 & $+0.061$ & $+0.008$ & 19 / 9  & $<$0.001 \\
    BA, whole test  & 0.005 & $+0.023$ & $+0.004$ & 15 / 7  & $<$0.001 \\
    Worst-group BA  & 0.019 & $+0.038$ & $+0.014$ & 2 / 0   & 0.014 \\
    \midrule
    \multicolumn{6}{@{}l}{\emph{Multilingual toxicity, 20 contrasts}} \\
    Worst class     & 0.018 & $+0.021$ & $+0.012$ & 0 / 10  & 0.161 \\
    BA, whole test  & 0.008 & $+0.011$ & $+0.006$ & 0 / 2   & 0.141 \\
    Worst-group BA  & 0.030 & $+0.034$ & $+0.019$ & 0 / 0   & 0.134 \\
    \bottomrule
  \end{tabular}
\end{table}

\paragraph{The WILDS metric.}
Worst-group accuracy as defined in WILDS takes the minimum of plain accuracy over the 16 cells formed
by the eight identities and the two labels. For the seed it is 0.530, in the cell of non-toxic
comments mentioning black people. Over the pool the minimum falls in that cell for 16 prompts, in the
cell of non-toxic comments mentioning white people for 16, and in two cells of toxic comments for the
remaining four. With a mean interval width of 0.079, the metric separates 19 of the 36 contrasts, and
the 11 that survive Holm correction are all losses. The best gain, $+0.065$, lies between the chance
baseline ($+0.042$) and the noise ceiling ($+0.080$), with a Reality Check $p$ of 0.146 (0.109 when
the contrasts are centred at their bootstrap means). Because the minimum sits mostly in non-toxic
cells, where errors are false alarms, the metric follows the share of comments the model labels toxic
(Spearman $\rho = -0.90$ across the 36 prompts). A large gain on one cell can be bought with recall on
the same identity: an APE final that raises accuracy on non-toxic comments mentioning black people by
$+0.22$ lowers recall on toxic ones from 0.824 to 0.608, and the balanced accuracy of that identity
barely moves (0.677 to 0.679). Our cells assign each comment to one identity, so these numbers apply
the WILDS metric to our test set rather than reproduce the WILDS benchmark.

\begin{figure}[H]
  \centering
  \includegraphics[width=0.72\linewidth]{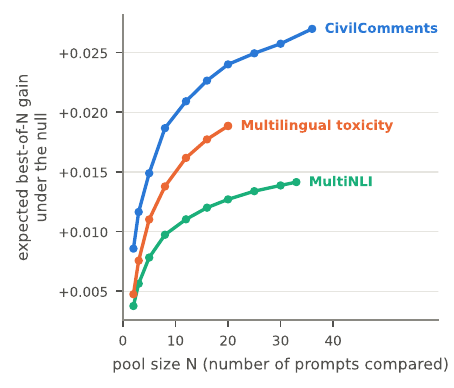}
  \caption{\textbf{The gain that noise alone produces grows with the number of prompts compared.}
    Expected best-of-$N$ gain in worst-group balanced accuracy under the null, for random subsets of
    each pool. With intervals only about a quarter wider, the curve for CivilComments ends 1.9 times
    as high as the one for MultiNLI, because its worst group changes between bootstrap resamples.}
  \label{fig:ceiling-vs-n}
\end{figure}

\paragraph{The minimum under resampling.}
On CivilComments the median prompt keeps its observed worst group in 47\% of bootstrap resamples, with
a range of 32--77\% over the prompts of the pool; for the seed the figure is 44\%. Among the 37
prompts, seed included, the worst group is \emph{white} for 26, \emph{black} and LGBTQ for five each
and \emph{muslim} for one. On MultiNLI the median is 91\% (63--96\%) and \emph{verbatim} is the worst genre
for all 34 prompts; on multilingual toxicity it is 93\% (59--100\%), with Amharic the minimum for 18
prompts and German for three. Figure~\ref{fig:ceiling-vs-n} shows how the chance baseline grows with
the number of prompts on each benchmark.

\paragraph{Smoothed minima.}
CVaR@25\% is the mean of the $\max(1, \lfloor K/4 \rfloor)$ lowest $\widehat{\mathrm{BA}}_g$. The
softmin first mixes each $\widehat{\mathrm{BA}}_g$ with $\kappa$ pseudo-examples at the mean over
groups $\overline{\mathrm{BA}}$, $\widetilde{\mathrm{BA}}_g = (n_g \widehat{\mathrm{BA}}_g + \kappa\,
\overline{\mathrm{BA}}) / (n_g + \kappa)$, where $n_g$ is the number of examples of group $g$, and
returns $-\tau \log \frac1K \sum_g \exp(-\widetilde{\mathrm{BA}}_g / \tau)$ with $\tau = 0.1$. With
$\kappa = 0$ there is no shrinkage; Table~\ref{tab:stable} covers $\kappa$ from 0 to 320.

\paragraph{Replicability.}
For Table~\ref{tab:stable} each test set is split into two halves 100 times, with 100 examples per
group--label cell in a half on CivilComments and 125 on the other two benchmarks. On each half every
pair of prompts among the seed and the optimizer finals is compared with the paired bootstrap, and a difference counts as distinguishable when its 95\% interval excludes zero. The minimum yields the fewest distinguishable differences, 0.5\%, 5.3\% and 10.7\% of the pairs on CivilComments, MultiNLI and multilingual
toxicity, and none of them is distinguishable with the opposite sign on the other half. The smoothed statistics rank prompts almost as the mean does: on CivilComments their rank
correlation with mean balanced accuracy is 0.96--0.99, and with the minimum 0.60--0.69. On MultiNLI,
CVaR@25\% both replicates (92.9\%) and agrees with the minimum ($\rho = 0.94$), because there the
prompts move all genres in the same direction; the minimum itself correlates with the mean at 0.82
on MultiNLI and at 0.54 on CivilComments.

\begin{table}[H]
  \caption{\textbf{How stable worst-group statistics are.} Upper part: share of differences
    distinguishable on one half of the test that have the same sign on the other half, over all pairs
    among the seed and the optimizer finals and 100 random halvings; the softmin range covers
    shrinkage toward the mean over groups with weights $\kappa$ from 0 to 320 examples, and with six
    groups CVaR@25\% equals the minimum. Lower part: medians over the prompts of the pool, 4,000
    bootstrap resamples.}
  \label{tab:stable}
  \centering
  \small
  \setlength{\tabcolsep}{3pt}
  \begin{tabular}{@{}lccc@{}}
    \toprule
     & \begin{tabular}[b]{@{}c@{}}Civil-\\Comments\end{tabular}
     & MultiNLI
     & \begin{tabular}[b]{@{}c@{}}Multilingual\\toxicity\end{tabular} \\
    \midrule
    \multicolumn{4}{@{}l}{\emph{Distinguishable differences that replicate, \%}} \\
    Worst-group BA $W$ & 60.3 & 82.0 & 86.5 \\
    CVaR@25\%          & 68.5 & 92.9 & = min \\
    Softmin            & 86.8--89.1 & 97.1--97.2 & 91.6--94.0 \\
    Mean BA            & 89.0 & 97.1 & 94.8 \\
    Worst class        & 99.2 & 98.4 & 95.6 \\
    \midrule
    \multicolumn{4}{@{}l}{\emph{The minimum under resampling}} \\
    Stays the worst, \% & 47 & 91 & 93 \\
    Downward bias      & 0.0131 & 0.0012 & 0.0006 \\
    Groups as minimum  & 4 of 8 & 1 of 10 & 2 of 6 \\
    \bottomrule
  \end{tabular}
\end{table}

\section{The Hidden Gain on Multilingual Toxicity}
\label{app:mechanism}

\paragraph{Per-language accuracy.}
Table~\ref{tab:per-language} gives balanced accuracy per language for the seed and the two GEPA finals.
Both finals raise Amharic and lower German, which becomes the minimum. Table~\ref{tab:recall} splits
the change for the seed-43 final into recall and specificity. On Amharic, recall rises from 0.599 to
0.850 and specificity falls from 0.843 to 0.714. On German, recall falls from 0.795 to 0.687 while
specificity rises from 0.759 to 0.803.

\begin{table}[!htbp]
  \caption{\textbf{Balanced accuracy per language} on multilingual toxicity, main task model. Bold
    marks the minimum, which moves from Amharic to German.}
  \label{tab:per-language}
  \centering
  \small
  \setlength{\tabcolsep}{3pt}
  \begin{tabular}{@{}lcccccc@{}}
    \toprule
    Prompt & en & de & ru & ar & hi & am \\
    \midrule
    Seed            & 0.921 & 0.777 & 0.972 & 0.799 & 0.819 & \textbf{0.721} \\
    GEPA, seed 43   & 0.939 & \textbf{0.745} & 0.960 & 0.791 & 0.857 & 0.782 \\
    GEPA, seed 42   & 0.942 & \textbf{0.755} & 0.968 & 0.766 & 0.845 & 0.776 \\
    \bottomrule
  \end{tabular}
\end{table}

\begin{table}[!htbp]
  \caption{\textbf{Recall (TPR) and specificity (TNR) per language} for the seed and the GEPA
    final of seed 43.}
  \label{tab:recall}
  \centering
  \small
  \setlength{\tabcolsep}{3pt}
  \begin{tabular}{@{}lcccccc@{}}
    \toprule
     & \multicolumn{3}{c}{Seed} & \multicolumn{3}{c}{GEPA, seed 43} \\
    \cmidrule(lr){2-4} \cmidrule(l){5-7}
    Language & TPR & TNR & BA & TPR & TNR & BA \\
    \midrule
    en & 0.980 & 0.862 & 0.921 & 0.970 & 0.908 & 0.939 \\
    de & 0.795 & 0.759 & 0.777 & 0.687 & 0.803 & 0.745 \\
    ru & 0.984 & 0.960 & 0.972 & 0.952 & 0.968 & 0.960 \\
    ar & 0.851 & 0.748 & 0.799 & 0.754 & 0.828 & 0.791 \\
    hi & 0.665 & 0.972 & 0.819 & 0.786 & 0.928 & 0.857 \\
    am & 0.599 & 0.843 & 0.721 & 0.850 & 0.714 & 0.782 \\
    \bottomrule
  \end{tabular}
\end{table}

\paragraph{All nine finals on the initially worst group.}
Table~\ref{tab:amharic} lists the gain on the initially worst group of every toxicity final. Only the two GEPA
finals are significant after Holm correction over the nine. Both remain significant when the
bootstrap is repeated with 100,000 resamples under three random seeds, and when unparsable examples
are dropped only over the nine finals and the seed instead of the whole pool. On German, the two
GEPA finals lose $-0.032$ $[-0.062, -0.002]$ and $-0.022$ $[-0.052, +0.008]$; neither loss survives a
correction over the six languages.

\begin{table}[!htbp]
  \caption{\textbf{Gain on the initially worst group for the nine toxicity finals} on the main task model.}
  \label{tab:amharic}
  \centering
  \small
  \setlength{\tabcolsep}{3pt}
  \begin{tabular}{@{}lcc@{}}
    \toprule
    Final & Gain on Amharic [95\% CI] & Holm \\
    \midrule
    GEPA, seed 43         & $+0.061$ $[+0.023, +0.101]$ & yes \\
    GEPA, seed 42         & $+0.055$ $[+0.019, +0.095]$ & yes \\
    APE, seed 42          & $+0.016$ $[-0.008, +0.040]$ & no \\
    APE, seed 43          & $+0.014$ $[-0.014, +0.043]$ & no \\
    ProTeGi, seed 44      & $+0.010$ $[-0.018, +0.039]$ & no \\
    EvoPrompt-DE, seed 44 & $+0.008$ $[-0.026, +0.045]$ & no \\
    EvoPrompt-DE, seed 42 & $-0.004$ $[-0.034, +0.025]$ & no \\
    ProTeGi, seed 43      & $-0.006$ $[-0.034, +0.022]$ & no \\
    ProTeGi, seed 42      & $-0.030$ $[-0.063, +0.002]$ & no \\
    \bottomrule
  \end{tabular}
\end{table}

\begin{figure}[H]
  \centering
  \includegraphics[width=\linewidth]{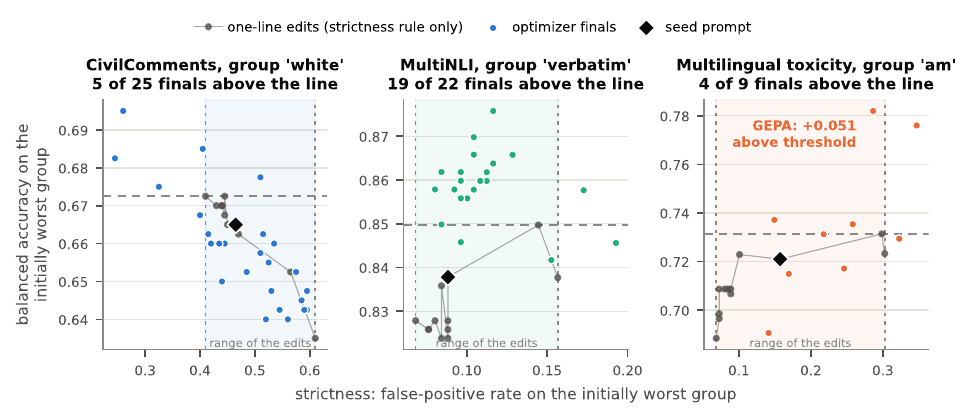}
  \caption{\textbf{What a threshold shift alone achieves on the initially worst group, and where the finals lie.} False-positive rate ($x$) and balanced accuracy ($y$) on $g_0$. Grey points are the
    one-line edits, joined in order of false-positive rate only to guide the eye; the dashed line
    marks the best of them, and the shaded band the range of false-positive rates they cover. A final
    above the dashed line improved discrimination beyond a threshold shift; outside the band the
    comparison is an extrapolation. Panel titles count the finals above the line on $g_0$.}
  \label{fig:strictness}
\end{figure}

\paragraph{Threshold sweep.}
Table~\ref{tab:sweep} gives the operating points behind Figure~\ref{fig:strictness} for Amharic.

 The
eight edits that state a probability threshold all make the model more lenient than the seed on
Amharic and stay between 0.688 and 0.723. The edit that asks the model to prefer the toxic label when
unsure reaches the highest balanced accuracy among the edits, 0.731. The seed-43 GEPA final lies inside the
range of false-positive rates covered by the edits and 0.051 above the best of them. Across the three
benchmarks, the edits cover false-positive rates of 0.41--0.61 on CivilComments, 0.068--0.157 on
MultiNLI and 0.069--0.302 on multilingual toxicity, and the best edit reaches 0.673, 0.850 and 0.731 on the initially worst group.
At almost the same false-positive rate, 0.29 against 0.30, the seed-43 final catches 85\% of the
toxic Amharic comments and the best edit 76\%. The seed-42 final lies outside the range of the
edits, at a false-positive rate of 0.35. In Figure~\ref{fig:strictness}, 19 of the 22 MultiNLI finals
lie above the best edit on the initially worst group, nearly all inside the range of the edits; on
CivilComments five of the 25 do, four of them outside the range, and none of their gains on that group is significant. Counted by worst-group balanced accuracy instead of the initially worst group, two of the 25
CivilComments finals lie above the line.

\begin{table}[!htbp]
  \caption{\textbf{Operating points on Amharic}, main task model. The edits are those of
    Table~\ref{tab:edits}.}
  \label{tab:sweep}
  \centering
  \footnotesize
  \setlength{\tabcolsep}{2pt}
  \begin{tabular}{@{}lccc@{}}
    \toprule
    Prompt & FPR & TPR & BA \\
    \midrule
    Lenient edit                & 0.089 & 0.502 & 0.707 \\
    Thresholds 20--90\%       & 0.069--0.101 & 0.445--0.547 & 0.688--0.723 \\
    Seed                        & 0.157 & 0.599 & 0.721 \\
    Strict edit                 & 0.298 & 0.761 & 0.731 \\
    Strictest edit              & 0.302 & 0.749 & 0.723 \\
    GEPA, seed 43               & 0.286 & 0.850 & 0.782 \\
    GEPA, seed 42               & 0.347 & 0.899 & 0.776 \\
    \bottomrule
  \end{tabular}
\end{table}

\section{Protocol Sensitivity}
\label{app:protocol}

\paragraph{Selection protocols.}
A selection protocol decides which of the prompts found by a search is reported. We vary four of its
parts: how the validation budget is spread over cells (uniformly, or a pilot followed by the three
worst groups of the pilot), the statistic used to pick the winner on validation data (six options,
from worst-group balanced accuracy to overall accuracy), a length penalty of
$0.00025 \cdot \max(0, \text{words} - 300)$ or none, and the rule for ties (first, random or
shortest). This gives 72 protocols. Each protocol is
applied to the finished pool: the validation examples are drawn 200 times, the protocol picks a
winner on each draw, and the winner is compared with the seed on the test set. Table~\ref{tab:protocols}
shows how the outcome varies across protocols.

\begin{table}[!htbp]
  \caption{\textbf{Outcomes across the 72 selection protocols.} Share of validation draws in which
    the selected prompt beats the seed on the test set, by point estimate and significantly after Holm
    correction over the pool; median over protocols, with the range in parentheses.}
  \label{tab:protocols}
  \centering
  \small
  \setlength{\tabcolsep}{4pt}
  \begin{tabular}{@{}llcc@{}}
    \toprule
    Benchmark & Metric & Point wins over the seed, \% & Significant wins after Holm, \% \\
    \midrule
    CivilComments & worst-group BA & 42.7 (15.5--86.0) & 0.0 (0--0) \\
    CivilComments & CVaR@25\%      & 47.5 (21.5--88.5) & 0.0 (0--0) \\
    MultiNLI      & worst-group BA & 100 (95--100)     & 18.0 (0--93.5) \\
    MultiNLI      & CVaR@25\%      & 100 (95.5--100)   & 97.0 (62--100) \\
    \bottomrule
  \end{tabular}
\end{table}

On CivilComments no protocol yields a significant win; the protocol changes only how often the
selected prompt looks better by point estimate. On MultiNLI, by worst-group balanced accuracy, the
share of significant wins ranges from none to 93.5\% depending on the protocol. Decomposing this
variation over the four parts of the protocol, the length penalty accounts for 92.2\% of it, the
validation budget for 2.1\%, the selection statistic for 1.6\% and the tie rule for none. Without the
penalty the share of significant wins is 75\% on average over 36 protocols; with it, it is zero for
all 36. The penalty takes 0.0635 from the best final of the pool, a GEPA prompt of 554 words, which is more
than the spread of the whole pool in worst-group balanced accuracy, 0.0519. By CVaR@25\% the selection statistic matters most (47.2\% of the variation), and on
CivilComments it does too (36--49\% of the variation in point wins).

\paragraph{Groupings with a meaning.}
Table~\ref{tab:supergroups} repeats the analysis of Section~\ref{sec:winner} for coarser groupings
built from existing categories. On CivilComments no grouping yields a significant gain. On MultiNLI
every grouping does, and coarser groupings give more findings. On CivilComments coarser groups also
dilute the effect: the mean absolute contrast over the pool is 0.0119 with eight identities, 0.0074
with three supergroups and 0.0062 with two groups.

\begin{table}[!htbp]
  \caption{\textbf{Worst-group balanced accuracy on groupings with a meaning.} Best gain, contrasts
    significant after Holm correction and Reality Check $p$.}
  \label{tab:supergroups}
  \centering
  \small
  \setlength{\tabcolsep}{3pt}
  \begin{tabular}{@{}lcccc@{}}
    \toprule
    Grouping & $k$ & \begin{tabular}[b]{@{}c@{}}Best\\gain\end{tabular} & Holm & $p$ \\
    \midrule
    \multicolumn{5}{@{}l}{\emph{CivilComments, 36 contrasts}} \\
    Identities              & 8  & $+0.019$ & 0  & 0.743 \\
    Identity supergroups    & 3  & $+0.014$ & 0  & 0.561 \\
    Religion vs.\ rest      & 2  & $+0.013$ & 0  & 0.271 \\
    \midrule
    \multicolumn{5}{@{}l}{\emph{MultiNLI, 33 contrasts}} \\
    Genres                  & 10 & $+0.038$ & 2  & 0.014 \\
    Registers               & 5  & $+0.027$ & 2  & 0.015 \\
    Spoken vs.\ written     & 2  & $+0.035$ & 9  & 0.0003 \\
    Matched vs.\ mismatched & 2  & $+0.020$ & 12 & 0.002 \\
    \bottomrule
  \end{tabular}
\end{table}

\begin{figure}[H]
  \centering
  \includegraphics[width=0.70\linewidth]{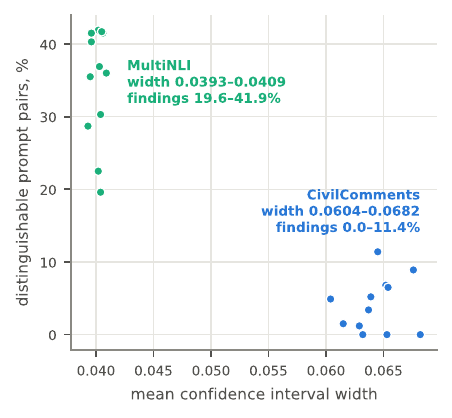}
  \caption{\textbf{Across test halves, interval width barely changes while the share of findings
    does.} Each point is one of 12 independent random halves of the test, all of the same size. The
    $x$-axis is the mean width of the 95\% intervals over all prompt pairs (CVaR@25\%), the $y$-axis
    the share of pairs whose interval excludes zero. On CivilComments none of the 86 pairs
    distinguishable on one half reversed sign on the other, so what varies is which differences
    become distinguishable.}
  \label{fig:draw-lottery}
\end{figure}

\paragraph{The test draw.}
We drew 12 independent random halves of each test set, all of the same size, and compared all pairs
of prompts on each half by CVaR@25\% (Figure~\ref{fig:draw-lottery}).

The mean interval width barely changes between
halves: its maximum is 1.13 times its minimum on CivilComments and 1.04 times on MultiNLI. The share
of pairs whose interval excludes zero varies from 0 to 11.4\% on CivilComments and from 19.6 to
41.9\% on MultiNLI. In 20 further splits of CivilComments into two halves, of the 86 pairs
distinguishable on one half, 64 had the same sign on the other half, 22 were not distinguishable
there, and none was distinguishable with the opposite sign.

\paragraph{The composition of the pool.}
The size of these effects depends on which prompts are in the pool. On ten random subpools of 15 of
the 37 CivilComments prompts, by CVaR@25\%, the median share of point wins over protocols ranges from
18\% to 50\%,
and the range of point wins across protocols from 48 to 78 percentage points.

\section{Live Search and the Ranking of Methods}
\label{app:live}

The protocol analysis of Appendix~\ref{app:protocol} selects among prompts that were already found.
In a live search the selection statistic also steers which prompts are found. We therefore ran live
optimization loops on CivilComments and MultiNLI: APE, EvoPrompt-DE and GEPA, each with three
selection statistics (worst-group balanced accuracy, its smoothed version and overall accuracy) and
three seeds, 27 runs per benchmark. On CivilComments 8 runs returned the seed, on MultiNLI 5, all of them
GEPA runs. The loops on CivilComments are separate from the pool of Section~\ref{sec:setup}.

\paragraph{Finals against the seed.}
Table~\ref{tab:live} compares the finals of the live loops with the seed. The picture of the fixed
pools repeats. On CivilComments no final wins significantly by a group metric or by overall
accuracy, while six win by worst-class accuracy. On MultiNLI two finals win by worst-group balanced accuracy and eleven by
CVaR@25\%.

\begin{table}[!htbp]
  \caption{\textbf{Live loops: finals that beat the seed significantly after Holm correction}, out of
    the finals that differ from the seed (in parentheses).}
  \label{tab:live}
  \centering
  \small
  \setlength{\tabcolsep}{4pt}
  \begin{tabular}{@{}lcc@{}}
    \toprule
    Metric & CivilComments (19) & MultiNLI (22) \\
    \midrule
    Worst-group BA  & 0 & 2 \\
    CVaR@25\%       & 0 & 11 \\
    Mean BA         & 0 & 17 \\
    Worst class     & 6 & 17 \\
    Overall accuracy & 0 & 17 \\
    \bottomrule
  \end{tabular}
\end{table}

\paragraph{What decides the quality of the final.}
We decomposed the variation of the finals' CVaR@25\% on the test set over seed, method and selection
statistic, with permutation tests within seeds. The method explains 20.2\% of the variation on
CivilComments ($p = 0.044$) and 52.4\% on MultiNLI ($p = 0.0012$); the selection statistic explains
8.1\% ($p = 0.34$) and 1.3\% ($p = 0.90$). On the fixed pool of MultiNLI the protocol mattered more
than the method: the standard deviation of outcomes across protocols within one study was 1.35--1.98
times that across methods; in the live loops the ratio is
0.64 on CivilComments and 0.16 on MultiNLI, with wide intervals. The selection statistic mainly
decides which candidate is reported. Re-selecting each run's archive with another statistic changes
the winner in 13 of 27 runs on CivilComments and 14 of 27 on MultiNLI, and changes the test quality of
the winner by $-0.0008$ $[-0.0093, +0.0060]$ and $+0.0034$ $[+0.0003, +0.0066]$ on average.

\paragraph{The ranking of methods.}
Table~\ref{tab:ranking} compares the three methods across benchmarks. APE beats the seed in all nine
MultiNLI runs, each significant after Holm correction, and in none of the nine CivilComments runs. On
multilingual toxicity, where the loops used ProTeGi as a fourth method and worst-group balanced
accuracy as the only selection statistic, GEPA has the largest mean gain. The ranking is a statement
about these configurations; the budgets are listed in Appendix~\ref{app:pool}.

\begin{table}[H]
  \caption{\textbf{The ranking of methods does not carry over.} MultiNLI and CivilComments: number of
    the nine runs whose final beats the seed by CVaR@25\% by point estimate, with a 95\% interval
    excluding zero, and after Holm correction. Toxicity: mean gain in worst-group balanced accuracy
    over three seeds.}
  \label{tab:ranking}
  \centering
  \small
  \setlength{\tabcolsep}{3pt}
  \begin{tabular}{@{}lccc@{}}
    \toprule
    Method & MultiNLI & CivilComments & Toxicity \\
    \midrule
    APE          & 9 / 9 / 9 & 0 / 0 / 0 & $+0.010$ \\
    EvoPrompt-DE & 9 / 5 / 1 & 5 / 0 / 0 & $-0.001$ \\
    GEPA         & 3 / 1 / 1 & 1 / 0 / 0 & $+0.019$ \\
    ProTeGi      & ---       & ---       & $-0.009$ \\
    \bottomrule
  \end{tabular}
\end{table}

\section{Selection Inside the Optimizer}
\label{app:validation}

A null result can have three causes: noise hides a gain, the optimizer's own selection on validation
data throws a gain away, or the search produces no distinguishable improvement. This appendix tests
the second on multilingual toxicity. For each of the 12 live runs, the final the method returned is
compared with every other candidate of the same run's archive on the first 1,000 test examples. The
archives hold 152 candidates, 138 of them distinct prompts. We report three estimates of the regret, the amount by which the best candidate
of the archive beats the final:
\begin{itemize}
  \item \emph{naive}: the best candidate on the test data minus the final. It is biased upward,
    because the best candidate is itself picked from 5 to 17 noisy estimates;
  \item \emph{selection-adjusted}: a Reality Check within each run---the maximum of the centred
    differences ``candidate minus final'' over the candidates, under the same paired bootstrap---with
    Holm correction across the 12 runs;
  \item \emph{cross-fit}: the best candidate is picked on one half of the test and compared with the
    final on the other, averaged over 200 stratified splits and both directions. This estimate
    carries no selection bias.
\end{itemize}

\begin{table}[!htbp]
  \caption{\textbf{Regret of the optimizer's own selection}, multilingual toxicity, 12 runs,
    worst-group balanced accuracy. Rows: unparsable answers dropped within the run, or scored as
    wrong. The interval is a bootstrap over runs. Under both rules no run is significant after Holm
    correction.}
  \label{tab:validation-summary}
  \centering
  \small
  \setlength{\tabcolsep}{3pt}
  \begin{tabular}{@{}lccc@{}}
    \toprule
    Unparsable & Naive & Cross-fit & 95\% CI \\
    \midrule
    dropped      & $+0.021$ & $-0.001$ & [$-0.011$, $+0.011$] \\
    scored wrong & $+0.017$ & $-0.001$ & [$-0.008$, $+0.008$] \\
    \bottomrule
  \end{tabular}
\end{table}

The naive estimate suggests that the archive held a prompt 0.021 better than the one returned. Once
the selection of the best candidate is accounted for, the regret is $-0.001$, and the interval excludes a regret
larger than 0.011 (Table~\ref{tab:validation-summary}). The candidates of one archive are
practically equivalent, so the selection has nothing to lose. No single run is significant
(Table~\ref{tab:validation-runs}). The largest case, ProTeGi with seed 42, has a naive regret of
$+0.068$ and a cross-fit regret of $+0.047$ at $p = 0.15$; a single run detects only a large regret,
since its own noise ceiling lies between 0.06 and 0.12.

\begin{table}[!htbp]
  \caption{\textbf{Regret per run}, worst-group balanced accuracy, unparsable answers dropped within
    the run. ``Kept'': candidates in the comparison, the final included. ``Excluded'': candidates
    whose inclusion would leave fewer than 700 examples parsable for all; the final is never
    excluded. ``Examples'': test examples left. ``Adjusted $p$'': selection-adjusted test; none
    survives Holm correction.}
  \label{tab:validation-runs}
  \centering
  \small
  \setlength{\tabcolsep}{4pt}
  \begin{tabular}{@{}lcccccccc@{}}
    \toprule
    Method & Seed & Kept & Excluded & Examples & Naive & Cross-fit & Adjusted $p$ & $C_{95}$ \\
    \midrule
    APE          & 42 & 7  & 0  & 997 & $+0.0119$ & $-0.0037$ & 0.764 & 0.063 \\
    APE          & 43 & 7  & 0  & 997 & $+0.0231$ & $-0.0004$ & 0.422 & 0.065 \\
    APE          & 44 & 7  & 0  & 997 & $+0.0007$ & $-0.0190$ & 0.856 & 0.063 \\
    ProTeGi      & 42 & 17 & 3  & 873 & $+0.0681$ & $+0.0469$ & 0.149 & 0.092 \\
    ProTeGi      & 43 & 17 & 1  & 899 & $0.0000$  & $-0.0051$ & 1.000 & 0.089 \\
    ProTeGi      & 44 & 17 & 2  & 816 & $0.0000$  & $-0.0304$ & 1.000 & 0.117 \\
    EvoPrompt-DE & 42 & 8  & 8  & 804 & $+0.0577$ & $+0.0211$ & 0.147 & 0.074 \\
    EvoPrompt-DE & 43 & 5  & 11 & 997 & $0.0000$  & $-0.0132$ & 1.000 & 0.060 \\
    EvoPrompt-DE & 44 & 7  & 9  & 969 & $+0.0170$ & $+0.0008$ & 0.688 & 0.077 \\
    GEPA         & 42 & 5  & 4  & 996 & $0.0000$  & $-0.0144$ & 1.000 & 0.077 \\
    GEPA         & 43 & 9  & 0  & 895 & $+0.0213$ & $-0.0063$ & 0.788 & 0.107 \\
    GEPA         & 44 & 6  & 2  & 898 & $+0.0543$ & $+0.0142$ & 0.216 & 0.092 \\
    \bottomrule
  \end{tabular}
\end{table}

\paragraph{Secondary metrics.}
For mean balanced accuracy the cross-fit regret is $+0.003$ [$-0.002$, $+0.009$]. For worst-class
accuracy it is $+0.024$ [$+0.007$, $+0.043$], but this is not a loss caused by noise: the runs select
by worst-group balanced accuracy, while the cross-fit picks by worst-class accuracy, and selecting by
one metric leaves room on another even without noise.

\paragraph{Candidates that break the answer format.}
Some candidates make the model answer in a form the parser cannot read
(Table~\ref{tab:broken}). They were also rejected on validation data, so no usable prompt was lost.
They do make the effective budgets unequal: half of the archive slots of EvoPrompt-DE are spent on
such candidates.

\begin{table}[!htbp]
  \caption{\textbf{Candidates that break the answer format}, share of unparsable answers on the first
    1,000 toxicity test examples.}
  \label{tab:broken}
  \centering
  \small
  \setlength{\tabcolsep}{4pt}
  \begin{tabular}{@{}lccc@{}}
    \toprule
    Method & \begin{tabular}[b]{@{}c@{}}Archive\\slots\end{tabular}
      & \begin{tabular}[b]{@{}c@{}}$\geq$50\%\\unparsable\end{tabular}
      & \begin{tabular}[b]{@{}c@{}}$\geq$95\%\\unparsable\end{tabular} \\
    \midrule
    EvoPrompt-DE & 48 & 25 & 21 \\
    GEPA         & 26 & 4  & 2 \\
    ProTeGi      & 57 & 2  & 1 \\
    APE          & 21 & 0  & 0 \\
    \bottomrule
  \end{tabular}
\end{table}

\paragraph{Deviation from the plan.}
The plan dropped an example with an unparsable answer for all candidates. Applied globally, this
removes every example, since 15 of the 138 candidates never produce a parsable answer. The rule was
therefore applied within each run, excluding a candidate whose inclusion would leave fewer than 700
examples, 40 candidates in all; the
decision was taken after a first look at the data and recorded in a separate amendment. Scoring
unparsable answers as wrong, which needs no exclusion, gives the same conclusion
(Table~\ref{tab:validation-summary}). Both preregistered predictions, that strong candidates are lost
in at least one run and that the loss matters on average, failed (Appendix~\ref{app:prereg}).

\paragraph{Scope.}
The test covers one benchmark and one task model, and it rules out only a regret larger than 0.011.

\section{The Second and Third Task Models}
\label{app:second-model}

\begin{table}[!htbp]
  \caption{\textbf{Gains on the initially worst group on the three task models}, with 95\% bootstrap intervals.
    $^{*}$Significant after Holm correction, over the nine finals on toxicity (finals that a model did
    not score enter with $p = 1$) and over the two planned contrasts on MultiNLI. On toxicity the main model uses the 2,907 test examples of the main
    analysis and the other two models their own valid examples (3,000 and 2,999); on MultiNLI all
    models use the same 5,000 examples. The last two columns give the gain on Gemma~3 12B minus the
    effect on the other model, both computed on the examples that all three models answered (2,996 on
    toxicity, 5,000 on MultiNLI) with one paired bootstrap; $^{\dagger}$significant after Holm
    correction over the eight differences.}
  \label{tab:transfer}
  \centering
  \footnotesize
  \setlength{\tabcolsep}{2.5pt}
  \begin{tabular}{@{}lccccc@{}}
    \toprule
    & \multicolumn{3}{c}{Gain on $g_0$} & \multicolumn{2}{c}{Gemma minus} \\
    \cmidrule(lr){2-4} \cmidrule(l){5-6}
    Final & Gemma~3 12B & GPT-4o-mini & Qwen3-235B & GPT & Qwen \\
    \midrule
    \multicolumn{6}{@{}l}{\emph{Multilingual toxicity, group $g_0$ Amharic}} \\
    GEPA, seed 43 & $+0.061$ $[+0.023, +0.101]^{*}$ & $-0.034$ $[-0.072, +0.002]$ & $-0.020$ $[-0.060, +0.016]$ & $+0.095^{\dagger}$ & $+0.080^{\dagger}$ \\
    GEPA, seed 42 & $+0.055$ $[+0.019, +0.095]^{*}$ & $-0.002$ $[-0.032, +0.030]$ & $-0.042$ $[-0.079, -0.006]$ & $+0.058^{\dagger}$ & $+0.098^{\dagger}$ \\
    \multicolumn{6}{@{}l}{\emph{MultiNLI, group $g_0$ verbatim}} \\
    APE, seed 43  & $+0.032$ $[+0.012, +0.052]^{*}$ & $-0.034$ $[-0.054, -0.016]^{*}$ & $-0.034$ $[-0.062, -0.008]^{*}$ & $+0.066^{\dagger}$ & $+0.066^{\dagger}$ \\
    GEPA, seed 42 & $+0.038$ $[+0.014, +0.062]^{*}$ & $-0.024$ $[-0.044, -0.006]^{*}$ & $+0.000$ $[-0.020, +0.020]$ & $+0.062^{\dagger}$ & $+0.038^{\dagger}$ \\
    \bottomrule
  \end{tabular}
\end{table}

\paragraph{The third model.}
Before scoring the finals we required a further model to be no more than 0.03 below the main model in
mean balanced accuracy for the seed prompt and no more than 0.05 below it on the initially worst group, with at
most 2\% unparsable answers. Qwen3-235B meets this on both benchmarks. Under the seed prompt its mean
balanced accuracy is 0.833 against 0.835 for the main model on multilingual toxicity and 0.933
against 0.897 on MultiNLI, and it orders the groups as the main model does (Spearman correlation 1.00
and 0.94). A smaller model of the same
family, Qwen3-30B, meets it on MultiNLI, where it also turns both gains negative on \emph{verbatim},
$-0.014$ $[-0.036, +0.008]$ and $-0.038$ $[-0.064, -0.010]$, the second significant after Holm
correction. On multilingual toxicity it reaches 0.623 on Amharic against 0.721 and is not used there.

\paragraph{The main and the second model.}
Table~\ref{tab:models} compares the main model with GPT-4o-mini under the seed prompt against the
same human labels. The second model is at least as accurate on average on all three benchmarks and on the initially worst group of MultiNLI.
Both models find the same groups hard: the rank correlation of the group order is high on toxicity
and CivilComments and moderate on MultiNLI, where the accuracies of the genres lie close together
(0.84--0.94). On toxicity and MultiNLI the answers of the two models agree least on the initially worst group.

\paragraph{Operating points on Amharic.}
The threshold argument of Section~\ref{sec:moves} reverses on the other two models. On GPT-4o-mini
the best of the one-line edits, scored on that model, reaches 0.716 on Amharic, and the two GEPA
finals fall below it, at 0.674 and 0.706. On Qwen3-235B both GEPA finals fall below the seed prompt
on Amharic, at 0.687 and 0.665 against 0.707.

\begin{table}[!htbp]
  \caption{\textbf{The main and the second task model under the seed prompt.} Balanced accuracy against the human
    labels, Spearman correlation of the group order, and the share of examples of the initially worst group on which the two models give the same answer (the lowest of all groups on MultiNLI and toxicity,
    third lowest of eight on CivilComments). On MultiNLI \emph{verbatim} is the second worst genre
    for the second model, 0.002 above \emph{slate}. On toxicity this table uses the 2,907 examples of
    the main analysis; Table~\ref{tab:transfer} and the operating points above use the second
    model's own 3,000 examples, on which its seed prompt reaches 0.708 on Amharic.}
  \label{tab:models}
  \centering
  \small
  \setlength{\tabcolsep}{3pt}
  \begin{tabular}{@{}lccc@{}}
    \toprule
     & \begin{tabular}[b]{@{}c@{}}Civil-\\Comments\end{tabular}
     & MultiNLI
     & \begin{tabular}[b]{@{}c@{}}Multilingual\\toxicity\end{tabular} \\
    \midrule
    Mean BA, main / second         & 0.717 / 0.721 & 0.897 / 0.908 & 0.835 / 0.838 \\
    Initially worst group, main / second    & 0.665 / 0.688 & 0.838 / 0.888 & 0.721 / 0.705 \\
    Worst group, main / second     & white / black & verbatim / slate & am / am \\
    Rank correlation of groups     & 0.74 & 0.55 & 1.00 \\
    Agreement on the initially worst group  & 85.2\% & 84.2\% & 72.1\% \\
    \bottomrule
  \end{tabular}
\end{table}

\paragraph{Reliability on CivilComments.}
A correlation between the effects of the same prompts on two models can only be as high as the
reliability of each measurement allows. We estimate split-half reliability from 500 random halvings
of the test within cells, corrected to the full test with the Spearman--Brown formula, for the
effects of the 25 CivilComments finals (Table~\ref{tab:reliability}). On the second model the effects on the initially worst group do not agree with themselves across halves, so the highest correlation that
could be observed is zero, and the observed $\rho = 0.04$ carries no information. Worst-class
accuracy, in contrast, is measured almost without noise by both models, and its effects agree only
weakly ($\rho = 0.28$): the prompts shift the balance between the classes differently on the two
models.

\paragraph{Replication of Sections~\ref{sec:winner} and~\ref{sec:stable}.}
On CivilComments the pool stays within chance on GPT-4o-mini as well: its best gain in worst-group
balanced accuracy is $+0.013$, below a chance baseline of $+0.024$ ($p = 0.81$). The minimum
replicates in 37\% of its distinguishable differences, against 82--87\% for the softmin family and 88\% for
the mean, and the observed worst group remains the worst in 52\% of bootstrap resamples.

\begin{table}[!htbp]
  \caption{\textbf{Reliability before correlation}, effects of the 25 CivilComments finals against
    the seed. Spearman correlation between the two models, split-half reliability on each, and the
    highest correlation the two reliabilities allow, $\sqrt{r_1 r_2}$.}
  \label{tab:reliability}
  \centering
  \small
  \setlength{\tabcolsep}{3pt}
  \begin{tabular}{@{}lcccc@{}}
    \toprule
    Effect & $\rho$ & \begin{tabular}[b]{@{}c@{}}Rel.\\main\end{tabular} & \begin{tabular}[b]{@{}c@{}}Rel.\\second\end{tabular} & Ceiling \\
    \midrule
    Initially worst group (white) & 0.04 & 0.38 & 0    & 0 \\
    Worst-group BA       & 0.26 & 0.33 & 0.08 & 0.16 \\
    Mean BA              & 0.16 & 0.80 & 0.63 & 0.71 \\
    Worst class          & 0.28 & 0.99 & 0.96 & 0.97 \\
    \bottomrule
  \end{tabular}
\end{table}

\section{PRIME}
\label{app:prime}

\paragraph{Method.}
PRIME alternates an evolutionary search over prompts with active selection of the training examples
the search learns from. In each of four cycles, an LLM mutator (DeepSeek-V4-Pro) rewrites a structured
prompt for 18 iterations with a population of 16. Every candidate is scored on a fixed set of 720
labelled training examples, 45 per group--label cell, by a smoothed minimum of the group balanced
accuracies, each shrunk toward the pooled mean; overall accuracy breaks near-ties, and prompts longer
than 300 words are penalized. Between cycles, PRIME selects the examples the mutator sees as feedback.
It starts from 24 labelled examples of a pool of 4,000 training examples and adds 12 per cycle,
choosing those on which three sampled answers of the task model disagree most. From these it builds a
batch of 24: 70\% are examples the current prompt gets wrong or is least certain about, with quotas
favouring the weakest groups, and the rest are correctly answered examples. Up to four examples from
the batch can enter the prompt as demonstrations. Prompts that are best on different groups are kept
on a Pareto front and seed the next cycle. A new prompt is accepted only if its smoothed minimum on 900
validation examples, 50 per cell, drops by at most 0.01, and the final prompt is chosen on these
validation examples.

\paragraph{Results on CivilComments.}
PRIME was run on CivilComments only, and its two finals are part of the CivilComments pool. Table~\ref{tab:methods-cc}
summarizes the finals of each method on that benchmark. PRIME has the only positive mean gain in
worst-group balanced accuracy and the best prompt of the pool, $+0.019$ $[-0.010, +0.039]$. No final
of any method is significant after Holm correction, and noise alone gives the best of 36 prompts at
least this gain in 74\% of resamples (Section~\ref{sec:winner}). On the initially worst group the same prompt
gains $+0.020$ ($p = 0.11$).

The comparison is also unequal in data. Most baselines drew on 360 labelled training examples
(Appendix~\ref{app:pool}). PRIME used 972 labelled training examples, 720 to score candidates, 180
for gate and audit checks and 72 acquired as feedback, and 900 labelled validation examples to choose
its final. Even a significant gain would therefore be confounded with access to labels.

\begin{table}[!htbp]
  \caption{\textbf{Methods on CivilComments}, gain in worst-group balanced accuracy over the seed for
    the 25 finals of the pool. For APE the mean is given per variant. No gain is significant after
    Holm correction.}
  \label{tab:methods-cc}
  \centering
  \footnotesize
  \setlength{\tabcolsep}{2pt}
  \begin{tabular}{@{}lccc@{}}
    \toprule
    Method & Finals & Mean gain & Best gain \\
    \midrule
    PRIME                      & 2 & $+0.0143$ & $+0.0194$ \\
    EvoPrompt-GA               & 3 & $-0.0050$ & $-0.0050$ \\
    EvoPrompt-DE               & 3 & $-0.0075$ & $-0.0025$ \\
    APE, 3 variants            & 7 & $-0.025$..$-0.008$ & $+0.0012$ \\
    ProTeGi, random labels     & 2 & $-0.0104$ & $-0.0033$ \\
    GPO                        & 3 & $-0.0119$ & $-0.0032$ \\
    ProTeGi                    & 3 & $-0.0258$ & $-0.0225$ \\
    GEPA                       & 1 & $-0.0325$ & $-0.0325$ \\
    ProTeGi, all labels        & 1 & $-0.0250$ & $-0.0250$ \\
    \bottomrule
  \end{tabular}
\end{table}

\section{Reproducibility and Cost}
\label{app:repro}

\paragraph{Cost.}
All model queries went through OpenRouter. The whole project, including runs not reported in this
paper, cost \$89.82. The second model on multilingual toxicity took 27,000 queries for the seed,
four finals and four edits, \$2.34 and 37 minutes; on the two MultiNLI finals and the seed, 15,000 queries and \$1.37. The third model took
42,000 queries and \$2.31 for the same contrasts on both benchmarks. The main
CivilComments matrix took about 18.5 hours of query time with four parallel streams.

\paragraph{Reproduction.}
The public repository in the Data Availability Statement contains saved predictions, split
indices, labels, group identifiers, and analysis code. Recomputing statistics on the saved
predictions requires no new queries or API key. The repository includes entry points for the
comparison with the seed and the Reality Check (Table~\ref{tab:winner},
Figure~\ref{fig:winners-curse}), the gain on the initially worst group and the operating points
(Section~\ref{sec:moves}, Appendix~\ref{app:mechanism}), replication and the stability of the
minimum (Table~\ref{tab:stable}), random and meaningful regroupings (Figure~\ref{fig:group-count},
Table~\ref{tab:supergroups}), the protocol simulation and the test draw (Appendix~\ref{app:protocol}),
the live loops (Appendix~\ref{app:live}), the selection study (Appendix~\ref{app:validation}), the
second and third models and the reliability analysis (Section~\ref{sec:model},
Appendix~\ref{app:second-model}),
and the figures. It also includes the whole-test controls (Table~\ref{tab:metrics-cc},
\texttt{analysis.whole\_test\_control}) and the direct contrasts between model effects
(Table~\ref{tab:transfer}, \texttt{analysis.cross\_model\_difference}).

\section{Can Measurability Be Predicted?}
\label{app:predict}

It would help to know before running an optimizer whether a benchmark can resolve a worst-group gain
at all. We examined three candidate predictors that can be computed before the search; none of them
works on our benchmarks.

\paragraph{The gap between the two worst groups.}
A large gap between the worst and the second worst group under the seed should make the choice of
the minimum stable and the gain easier to see. The data show no such relation. The matched and
mismatched split of MultiNLI has a gap of 0.0003 and gives 12 significant gains, the most of any
grouping we studied. Religion against the rest on CivilComments has a gap of 0.0648 and gives none.
The gap does predict what it measures by construction, the stability of the minimum, but not whether
a gain will be found.

\paragraph{The leverage of the threshold.}
The leverage is the mean absolute change in worst-class accuracy over the eleven one-line edits,
divided by the mean absolute change in worst-group balanced accuracy over the same edits. A high
leverage suggests that the prompts move the threshold rather than the groups. It does not separate our
benchmarks. On multilingual toxicity it is 2.51,\footnote{Computed before the live loops on 2,985
test examples, the value the preregistration was compared with; on the 2,907 examples of the main
analysis the leverage is 2.54 with real groups and 5.05 with random ones.} close to the 2.29 of
MultiNLI, yet the best worst-group gain is within noise on toxicity and above it on MultiNLI. Random
groups of the same size give a higher leverage on all three benchmarks, 5.97, 3.64 and 5.98 against
5.26, 2.29 and 2.51 with the real groups, so the leverage reflects how noisy the minimum is under a
given grouping.

\paragraph{A noise ceiling calibrated on edits.}
The one-line edits are cheap to evaluate, so the noise ceiling could be estimated on them before the
search and extrapolated to the expected number of finals. On multilingual toxicity this gave $+0.031$
for twelve prompts, while the ceiling computed on the nine distinct finals after the search is
$+0.044$, 1.42 times higher, even though it applies to fewer prompts. Table~\ref{tab:edits-finals}
shows the reason. Optimizer finals disagree with the seed on 1.6 to 2.2 times as many examples as the
edits do, so their paired intervals are 1.3 to 1.6 times wider, and the same holds on all three
benchmarks. A ceiling for the finals therefore has to be computed on the finals.

\begin{table}[!htbp]
  \caption{\textbf{Why a ceiling calibrated on edits is too low.} Share of test examples on which a
    prompt answers differently from the seed, averaged over prompts, and the mean half-width of the
    95\% interval of the worst-group contrast.}
  \label{tab:edits-finals}
  \centering
  \small
  \setlength{\tabcolsep}{3pt}
  \begin{tabular}{@{}lcccc@{}}
    \toprule
     & \multicolumn{2}{c}{Disagree with seed} & \multicolumn{2}{c}{Half-width} \\
    \cmidrule(lr){2-3} \cmidrule(l){4-5}
    Benchmark & Edits & Finals & Edits & Finals \\
    \midrule
    CivilComments & 4.0\% & 8.9\% & 0.0176 & 0.0285 \\
    MultiNLI      & 2.6\% & 4.2\% & 0.0160 & 0.0204 \\
    Toxicity      & 4.2\% & 8.8\% & 0.0258 & 0.0355 \\
    \bottomrule
  \end{tabular}
\end{table}

\bibliographystyle{unsrtnat}
\bibliography{references}

\end{document}